\documentclass[10pt,twocolumn,letterpaper]{article}

\usepackage[pagenumbers]{cvpr} % To force page numbers, e.g. for an arXiv version

\usepackage[accsupp]{axessibility}  % Improves PDF readability for those with disabilities.

\usepackage{duckuments}

\usepackage{array}
\usepackage{relsize}
\usepackage{graphicx}
\usepackage{booktabs}
\usepackage{multirow}
\usepackage{makecell}
\usepackage{pifont}  % 添加这个包
\usepackage{natbib}  % 确保有这个包用于 \citep

\usepackage{algorithm}
\usepackage{algpseudocode}
\usepackage{amsmath}
\usepackage{cuted}

\usepackage{multibib}
\usepackage{titletoc}

\usepackage[table,dvipsnames]{xcolor}  % 统一xcolor选项
\definecolor{mgray}{gray}{.9}
\definecolor{dgray}{gray}{.5}

\newcommand{\cmtt}[1]{{\fontfamily{cmtt}\selectfont #1}}

\newcommand{\set}[1]{\left\{#1\right\}}

\newcommand{\name}{SimScale}
\newcommand{\we}{data engine\xspace}
\newcommand{\boldparagraph}[1]{\vspace{0.1cm}\noindent{\bf #1}}

\newcommand{\tightbox}[2][0.9,0.95,1]{%
  \begingroup
  \setlength{\fboxsep}{2pt}% padding
  \colorbox[rgb]{#1}{#2}%
  \endgroup
}

\newcolumntype{P}[1]{>{\centering\arraybackslash}p{#1}}

\definecolor{cvprblue}{rgb}{0.21,0.49,0.74}
\usepackage[pagebackref,breaklinks,colorlinks,allcolors=cvprblue]{hyperref}

\title{\textsc{\name}:~Learning to Drive via Real-World Simulation at Scale}

\author{
Haochen Tian$^{1,2,3\diamond}$\quad
Tianyu Li$^{2\dagger}$\quad
Haochen Liu$^{3}$\quad
Jiazhi Yang$^{2}$\quad 
Yihang Qiu$^{2,3\diamond}$
\\
Guang Li$^{3}$ \quad
Junli Wang$^{1,3}$ \quad
Yinfeng Gao$^{1,3}$ \quad
Zhang Zhang$^{1\S}$ \quad
Liang Wang$^{1\S}$ \quad 
\\
Hangjun Ye$^{3}$ \quad
Tieniu Tan$^{1}$ \quad
Long Chen$^{3\S}$ \quad
Hongyang Li$^{2}$
\\[2mm]
$^1$ MAIS, Institute of Automation, Chinese Academy of Sciences\\
$^2$ OpenDriveLab at The University of Hong Kong\quad
$^3$ Xiaomi EV
\\[1.5mm]
{
\hypersetup{urlcolor=Violet}
\href{https://opendrivelab.com/\name}
{\texttt{{https://opendrivelab.com/\name}}}
}
}

\begin{document}

\twocolumn[{
\renewcommand\twocolumn[1][]{#1}
\maketitle
\vspace{-1cm}
\begin{center}
    \centering
    \captionsetup{type=figure}\includegraphics[width=\textwidth]{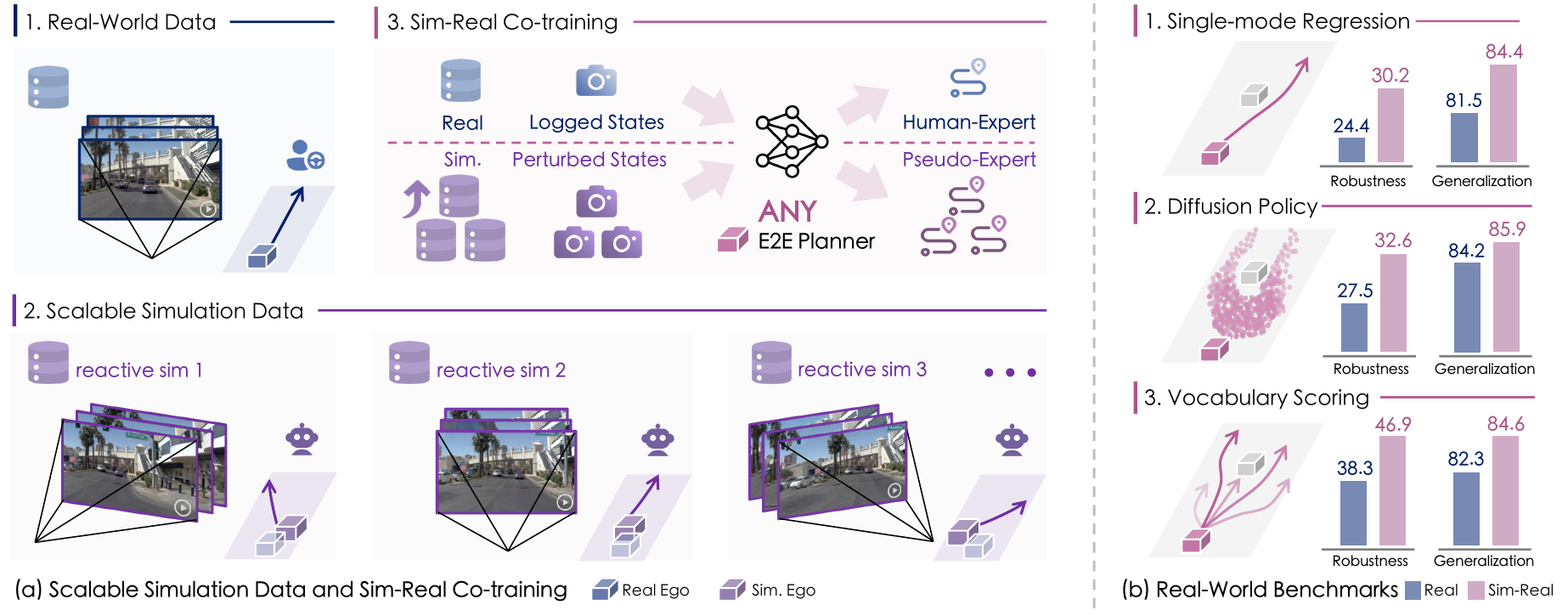} 
    \vspace{-0.6cm}
    \captionof{figure}{
    \textbf{Scaling up end-to-end planners by simulation.} 
    (a) We construct large-scale simulation data by perturbing ego trajectories, generating corresponding pseudo-expert demonstrations, and rendering multi-view observations in reactive environments. Combined with real-world data, this enables broad coverage of out-of-distribution states and supports sim–real co-training for \textit{any} end-to-end planner.
    (b) Across three representative planner families, including regression, diffusion, and vocabulary scoring, sim-real co-training consistently produces synergistic improvements in robustness and generalization, demonstrating clear and predictable simulation scaling trends.
}
\label{fig:teaser}
\end{center}}]

{\let\thefootnote \relax
\footnote{
\hangindent=1.8em
$^\diamond$ Work done while interning at Xiaomi Embodied Intelligence Team.\\
$^\dagger$ Project Lead. $^\S$ Equal Advising. \\
Primary contact to Haochen Tian \texttt{tianhaochen2023@ia.ac.cn}
}}

\vspace{-8pt}

\begin{abstract}

Achieving fully autonomous driving systems requires learning rational decisions in a wide span of scenarios, including safety-critical and out-of-distribution ones.
However, such cases are underrepresented in real-world corpus collected by human experts.
To complement for the lack of data diversity, we introduce a novel and scalable simulation framework capable of synthesizing massive unseen
states upon existing driving logs.
Our pipeline utilizes advanced neural rendering with a reactive environment to generate high-fidelity multi-view observations controlled by the perturbed ego trajectory.
Furthermore, we develop a pseudo-expert trajectory generation mechanism for these newly simulated states to provide action supervision.
Upon the synthesized data, we find that a simple co-training strategy on both real-world and simulated samples can lead to significant improvements in both robustness and generalization for various planning methods on challenging real-world benchmarks, up to +8.6 EPDMS on navhard and +2.9 on navtest.
More importantly, such policy improvement scales smoothly by increasing simulation data only, even without extra real-world data streaming in.
We further reveal several crucial findings of such a sim-real learning system, which we term \textbf{\name}, including the design of pseudo-experts and the scaling properties for different policy architectures.
Simulation data and code have been released at\hypersetup{urlcolor=Violet}
\href{https://github.com/OpenDriveLab/\name}
{\texttt{https://github.com/OpenDriveLab/\\\name}}.

\end{abstract}
\vspace{-12pt}

\section{Introduction}
\label{sec:intro}

Data scaling is recognized as a foundational principle in modern deep learning across various domains, including language, vision, and multimodal modeling, underpinning steady performance improvements as data sizes increase~\cite{kaplan2020scalinglaw,brown2020gpt3,zhai2022scalingvit,radford2021clip}. In autonomous driving, end-to-end (E2E) planning learns to map raw observations to actions, offering a promising way to leverage large-scale driving data to enable the emergence of fully autonomous systems~\cite{hwang2024emma,naumann2025e2escaling,zheng2024preliminaryscaling}.

Nevertheless, real-world driving data from human expert demonstrations are dominated by common scenarios, while non-trivial cases, \eg, safety-critical, are underrepresented~\cite{chen2023e2esurvey,xu2025wod,liu2025reinforced,yang2025resim}. 
Moreover, planners trained on such data are confined to human driving distribution and struggle to generalize to rare or unseen situations, leading to distribution shift and causal confusion in deployment~\cite{gao2025rad, li2025hydramdp}.
Consequently, scaling real-world data only is inefficient for achieving deploy-ready autonomous driving.

Simulation via neural rendering~\cite{kerbl2023_3dgs} can generate high-fidelity driving scenarios and thus has the potential to produce out-of-distribution (OOD) states deviated from human demonstrations at scale, which is essential for closed-loop planning~\cite{goff2025ldwm,popov2024mitigating}. Therefore, scaling simulation data presents an attractive alternative to solely relying on real-world data. 
However, planners require feasible corresponding demonstrations to learn how to handle OOD states, while current simulation methods fail to generate such demonstrations effectively. 
Moreover, the impact of scaling simulation data lacks in-depth analysis.
In this work, we aim to derive a systematic recipe for scaling simulation data from limited real-world scenarios in end-to-end planning.

To conduct comprehensive experiments and analyses, this study is structured along \textbf{three key questions}: 
\textit{(1) what constitutes effective simulation data, (2) how well planners benefit from it, and (3) whether this system scales predictably upon fixed real-world corpus.}

To this end, we formulate a scalable simulation data generation framework that extends the expert distribution from existing real-world training data to bootstrap end-to-end autonomous driving systems.
We develop a 3D Gaussian Splatting (3DGS)~\cite{kerbl2023_3dgs} simulation data engine, which allows controlling the temporal ego and other agent states and rendering multi-view videos from the ego’s perspective. 
Concretely, we first sample a diverse set of plausible perturbations to the ego trajectory, maximizing coverage of the state space, \eg, off-center lane drifts, close interactions, among others. 
Then, we take the final state of each perturbed trajectory as the perturbed state and generate corresponding demonstrations using pseudo-experts of two forms in comparison.
The first, recovery-based expert retrieves trajectories that steer the policy within the human trajectory manifold, resulting in human-like yet cautious behaviors.
The second, privileged planner-based expert~\cite{dauner2023pdmclosed} generates the trajectory that maximizes optimality, representing an exploratory strategy with lower realism. To enhance scalability and plausibility, the entire pipeline is executed in a reactive environment~\cite{treiber2000idm}, where surrounding agents interact with the ego vehicle responsively.

To thoroughly assess the effect of simulation data, we consider three types of end-to-end planners with various model scales, \ie, LTF~\cite{kashyap2022transfuser} for regression methods, DiffusionDrive~\cite{liao2025diffusiondrive} for diffusion-based planners, and GTRS-Dense~\cite{li2025gtrs} for vocabulary scoring.
We employ a simple yet effective sim-real co-training strategy~\cite{maddukuri2025simreal,nv2025gr00tn1} to maintain human driving distribution while mitigating visual domain degradation. Furthermore, by keeping a constant amount of real data and progressively adding simulation data via non-overlapping samples, we investigate how diverse planners can benefit from simulation data and the general scaling property.
We utilize two real-world closed-loop benchmarks to evaluate the planners from multiple perspectives. \cmtt{navhard}~\cite{Cao2025navsimv2} focuses on unseen, challenging scenarios to assess the impact of OOD states on planners, while \cmtt{navtest}~\cite{Dauner2024navsim} offers a broad set of diverse scenarios to test planners’ ability to handle varying situations. 

The complete sim-real learning system, which comprises a scalable simulation data construction pipeline and an effective sim–real co-training strategy as shown in Fig.~\ref{fig:teaser}, is termed $\textbf{\name}$. Rigorous experiments uncover crucial findings enabled by \name---including, but not limited to, the following:

\begin{itemize}
    \item Scalable simulation with pseudo-expert unlocks the inherent potential of available real-world driving data.
    \item Sim-real co-training 
    improves both robustness and generalization synergistically in diverse end-to-end planners.
    \item Exploratory expert and interaction environments improve the effectiveness of simulation data.
    \item Planners with multimodal modeling capabilities exhibit more encouraging data scaling properties.
\end{itemize}

\begin{figure*}[t!]
    \centering
    \includegraphics[width=1\linewidth]{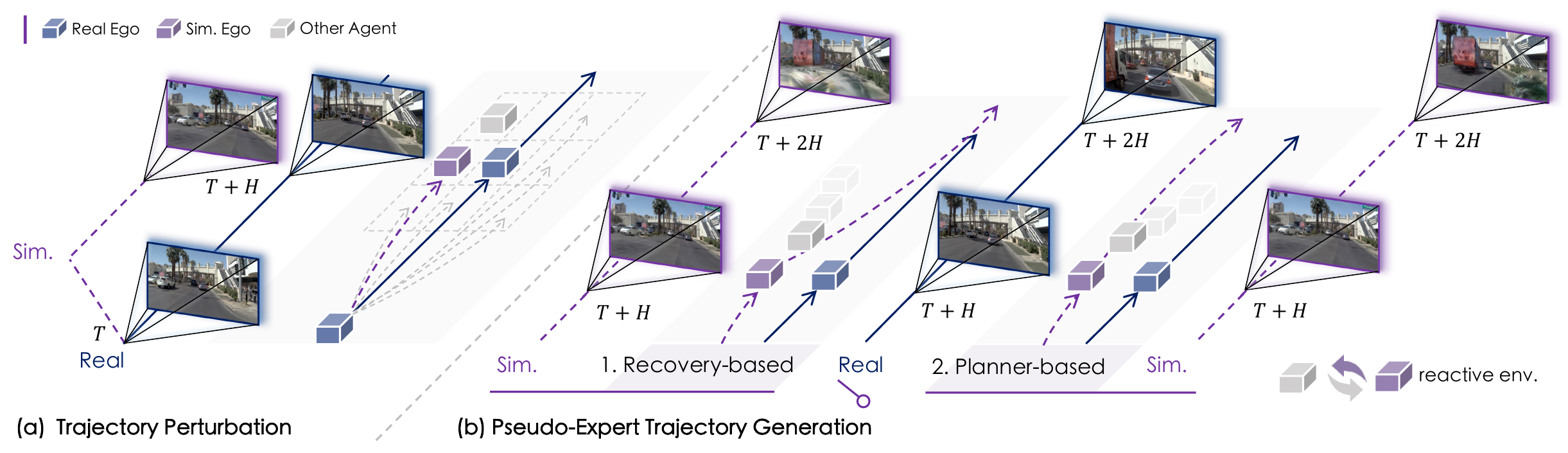}
    \caption{
    \textbf{Pseudo-expert scene simulation pipeline}. \textbf{(a)} Trajectory perturbation on $T$ to $T + H$, \textbf{(b)} reactive environment rollout, and pseudo-expert trajectory generation from $T+H$ to $T+2H$ under recovery-based and planner-based strategies.
    }
    \label{fig:pipeline}
    \vspace{-3pt}
\end{figure*}

\section{Methodology} \label{sec:method}

We outline \name~as follows. In Sec.~\ref {sec:engine}, we briefly introduce our 3GDS simulation data engine supporting controllable multi-view video rendering.
In Sec.~\ref {sec:pipeline}, we  propose a pseudo-expert scene
simulation pipeline to generate diverse simulation data containing OOD states with feasible
demonstrations.
In Sec.~\ref {sec:cotraining}, we demonstrate a scalable sim-real co-training method with different end-to-end planners.

\subsection{Preliminary}
End-to-end planning models take observations within a history window as input and output a predicted future trajectory. 
Each training scenario starts at a selected timestep and includes a history horizon of length $T$ and a planning horizon of length $H$. The model 
processes the past $T$ frames to predict the future $H$ frames, resulting in a complete training sample spanning $T+H$ timesteps.

\subsection{3DGS Simulation Data Engine}
\label{sec:engine}

To reduce the domain gap between real-world data and generated observations from novel views in simulated scenarios, a photorealistic \we is required. 
Built upon 3DGS~\cite{kerbl2023_3dgs} assets reconstructed from real-world datasets, our \we $\Phi(K_t, E_t,\set{x_{i,t},y_{i,t},\theta_{i,t}}_{i=1}^N)$ takes as input the camera intrinsics $K_t$ and extrinsics $E_t$ at timestep $t$, along with the positions and yaw angles $(x,y,\theta)$ of a non-ego vehicle $i$ at the same timestep, and renders the corresponding RGB observations from the camera. 
The camera extrinsics $E_t$ can easily be obtained from the ego vehicle's position and heading $(x_{0,t},y_{0,t},\theta_{0,t})$ with the ego-to-camera transformation, while the other camera parameters are directly adopted from the raw dataset.

\boldparagraph{Preprocessing.}
Following prior work~\cite{li2025mtgs}, we apply exposure alignment across images captured simultaneously by multi-view cameras at time $t$, using projected LiDAR points at $t$ as guidance. Moreover, with help of 3D bounding boxes provided by NAVSIM annotations, colored LiDAR points are divided into several groups, \ie, the static background and multiple vehicles, and further used as initialization for Gaussians to improve the reconstruction performance.

\boldparagraph{Block-wise Reconstruction.}
During scene reconstruction, increasing the number of images significantly raises computational cost and runtime. Thus, we perform reconstruction in a block-wise manner, where each block corresponds to a spatio-temporal range. 
Following previous work~\cite{yan2024streetgs, li2025mtgs}, backgrounds and foregrounds are reconstructed as separate models using the per-timestep locations and orientations of 3D bounding boxes, resulting in a static background asset and multiple movable vehicle assets. These assets could then be placed at desired locations and headings specified by inputs, rendering sensor data at novel views. 
Blocks with low average PSNR in novel view synthesis are excluded to ensure the quality of generated simulation data.

\subsection{Pseudo-Expert Scene Simulation}
\label{sec:pipeline}

Based on the data engine, we design a pseudo-expert scene simulation pipeline to generate diverse simulation data from existing real-world data, as shown in Alg.~\ref{algo:pipeline}.
The pipeline aims to produce feasible demonstrations from perturbed states paired with expert trajectories, 
as illustrated in Fig.~\ref{fig:pipeline}.

\boldparagraph{Reactive Scene Reconstruction.} For each training clip $d$, we perform two simulations of duration $H$: one for exploring perturbed states at $t = T$, and the other for generating expert trajectories at $t = T + H$.
In each simulation,  ego trajectory $\tilde{a}_{t:t+H}$ is simulated by LQR~\cite{lehtomaki2003lqr}, while other agents are modeled using the IDM~\cite{treiber2000idm} to interact with the ego, producing the corresponding future states $\tilde{s}_{t:t+H}$. For the valid expert trajectories, the simulated ego–agent trajectories cross two stage $(\tilde{a}_{T:T+2H}, \tilde{s}_{T:T+2H})$ are then rendered into multi-view videos $\tilde{o}_{T:T+2H}$ using the data engine $\Phi$.
This decoupling of behavior simulation from sensor rendering enables reactive environments where other agents respond plausibly to the ego's behaviors, thereby enhancing both the realism and diversity of the simulation data.

\boldparagraph{Trajectory Perturbation.} 
At the first simulation step ($t=T$), we perturb the ego trajectory $\tilde{a}_{t:t+H}$ so that the ego reaches a new terminal state at $t=T+H$, which then serves as the start state for the next rollout (Fig.~\ref{fig:pipeline} (a)). Our objective is to sample diverse yet plausible states. For diversity, perturbations are drawn from a clustered human-trajectory vocabulary that densely covers the action space. For plausibility, we restrict perturbations to remain near human behavior by thresholding longitudinal/lateral shifts $(r_{\mathrm{lon}},r_{\mathrm{lat}})$ and heading change $|\Delta\theta|$, and by removing trajectories that are physically invalid (collisions, off-road, unstable renderings). We further apply spatially sparse sampling on trajectory endpoints using an interleaved grid with steps $(\delta_{\mathrm{lon}},\delta_{\mathrm{lat}})$ to promote uniform coverage. Since reactive simulation is costly, we first filter infeasible trajectories non-reactively and run reactive checks only on the sparsed set. This produces a set of dynamically and physically feasible perturbations $a_{\mathrm{per}}\in\mathcal{A}_{\mathrm{per}}$.

\begin{algorithm}[t!]
\caption{Pseudo-Expert Scene Simulation}
\label{algo:pipeline}
\begin{algorithmic}[1]
\Require Training clip $d = (o_t, s_t, {a}_t)^{T+2H}_{t=0}\in\mathcal{D}$, perturb action sets $\mathcal{A}_\text{per}$, sensor engine $\Phi$, pseudo policy $\pi_\text{exp}$, reward functions $f$, reactive simulation $\mathcal{T}$.

\State Initialize simulation scene dataset $\mathcal{D}_{\text{sim}} \gets \emptyset$
\For{each perturbed action $a_\text{per}\in\mathcal{A}_\text{per}$}
  \For{timestep $t \in\{T, T+H\}$}

    \State \textbf{\texttt{1.Pseudo Action Generation:}}
        \If{$t=T$}  $\tilde{a}_{t:t+H} \gets a_\text{per};$ 
        \Else{ $\tilde{a}_{t:t+H} \gets \pi_\text{exp}(\tilde{o}_{t},\tilde{s}_{t})$.}
        \EndIf
    \State \textbf{\texttt{2.Reactive State Simulation:}} 
    \State $\tilde{s}_{t:t+H} \gets \mathcal{T}(\tilde{a}_{t:t+H},\tilde{s}_{t}).$
    \State \textbf{\texttt{3.Sensor Simulation:}}
    \State$\tilde{o}_{t:t+H} \gets \Phi(\tilde{s}_{t:t+H}).$
    \State \textbf{\texttt{4.Reward Filtering / Labeling:}} \State$\tilde{r}_{t:t+H} \gets f(\tilde{a}_{t:t+H},\tilde{s}_{t:t+H}).$
    \If{$\textbf{\texttt{Filtered}}(\tilde{r}_{t:t+H},\tilde{o}_{t:t+H})$}
        \State \texttt{Break}
    \EndIf
  \EndFor
    \State Append $(\tilde{o}_t, \tilde{s}_t, \tilde{a}_t, \tilde{r}_t)^{T+2H}_{t=0}$ to $\mathcal{D}_{\text{sim}}$  

\EndFor

\State \Return $\mathcal{D}_{\text{sim}}$
\end{algorithmic}
\end{algorithm}

\boldparagraph{Pseudo-Expert Trajectory Generation.} At the second simulation $t = T+H$, for each perturbed state, we employ a non-human expert, referring as the pseudo-expert $\pi_\text{exp}$, to generate a feasible corresponding trajectory $\tilde{a}_{t:t+H}$. Since the pseudo-expert is not perfect and $\tilde{a}_{t:t+H}$ will be used as supervision, we apply stricter filtering in the second stage simulation. In addition to physical constraints, traffic rules and vehicle kinematic limits are also enforced, following Eq.~\ref{eq:epdms}. To investigate which strategy best serves as supervision for end-to-end planners, we compared two strategies: a conservative \textit{recovery-based} expert, and an exploratory \textit{planner-based} expert, as detailed below. 

\boldparagraph{(1) Recovery-based Expert.} Recovery steers the policy within the human manifold after perturbation. To ensure robustness, our recovery expert $\pi_\text{exp}$ retrieves a human trajectory from a large vocabulary $\mathcal{V}_{\text{h}}$ that best matches the ego vehicle’s logged state at time $t = T+2H$ (Fig.~\ref{fig:pipeline} (b1)).
For each candidate trajectory of horizon $H$, we summarize its initial and final poses with a compact matching vector:
\begin{equation}
\mathbf{m} = [\tilde{v}^x_t, \tilde{v}^y_t,\tilde{\theta}_t, \tilde{x}_{t+H-1}, \tilde{y}_{t+H-1}, \tilde{\theta}_{t+H-1}].
\end{equation}
Given the ego’s perturbed state with target vector $\mathbf{m}_r$, $\pi_\text{exp}$ retrieves the closest human maneuver by:
\begin{equation}
\tilde{a}_{t:t+H} 
= \arg\min_{a \in \mathcal{V}_{\text{h}}} 
\|\mathbf{m}(a) - \mathbf{m}_{\text{r}}\|_1.
\end{equation}
This yields a human-like yet conservative fallback behavior, stabilizing under distributional drifts.

\boldparagraph{(2) Planner-based Expert.} 
Following prior works~\cite{zhang2021roach,beisswenger2024pdm}, 
we use a privileged planner $\mathbf{P}$ that leverages ground-truth states to generate reactive and optimized trajectory rollouts in simulation, as shown in Fig.~\ref{fig:pipeline} (b2).
The planner-based expert $\pi_\text{exp}\leftarrow\mathbf{P}$ is defined as:
$\tilde{a}_{t:t+H} 
= \mathbf{P}(\tilde{s}_{t:t+H}).$ Compared with the recovery policy, planner-based expert relies on rules or cost heuristics, trading human-likeness and realism of its behavior occasionally.
Still, it offers strong optimization and diverse exploratory rollouts, enriching expert supervision beyond human data.

\subsection{Scalable Sim-Real Co-training}
\label{sec:cotraining}
\boldparagraph{Strategies for Co-training.}
Sim-real co-training serves as a simple yet effective strategy~\cite{nv2025gr00tn1, maddukuri2025simreal,google2023rt2,yang2025resim} that enables the integration of both real and simulated data for planning. In our approach, we randomly sample from a mixture of real-world data $\mathcal{D}$ and simulation data $\mathcal{D}_\text{sim}$ during training, aiming to preserve the human driving distribution while mitigating visual domain degradation caused by potential simulation artifacts, such as subtle rendering inconsistencies, temporal jitter, or unrealistic lighting and shadows~\cite{liu2024enhance3dgs, liu2024novel}.
Our fully automated, scalable simulation data generation framework enables scaling the total training data by progressively adding non-overlapping simulation samples, while keeping the real data amount fixed.

\boldparagraph{Planners for Co-training.}
We aim to comprehensively evaluate the effectiveness of simulation data for end-to-end planners.
Modern end-to-end planning approaches can be broadly categorized into three representative paradigms: \textit{regression-based planners}~\cite{hu2023uniad,kashyap2022transfuser}, \textit{diffusion-based planners} ~\cite{liao2025diffusiondrive, xing2025goalflow}, and \textit{vocabulary scoring-based planners} ~\cite{chen2024vadv2,li2025hydramdp}.
Accordingly, we consider representative models from each paradigm in our co-training experiments.

\boldparagraph{(1) Co-training with Pseudo-Expert Trajectory.} 
Regression and diffusion-based planners rely on expert demonstrations. 
The co-training process can thus be formulated as:
\begin{equation}
\arg\min_{\theta}
\mathbb{E}_{(a,o)\sim (\mathcal{D}\,\cup\,\mathcal{D}_{\text{sim}})}
\Big[
  \mathcal{L}_{\text{im}}\big(a, \pi_\theta(\hat{a}|o)\big)
\Big],
\label{eq:im}
\end{equation}
here, $\mathcal{L}_{\text{im}}$ denotes the imitation loss; $\mathcal{D}$ and $\mathcal{D}_{\text{sim}}$ represent the real-world and generated simulation datasets, respectively; and $A$ refers to expert trajectories, i.e., human-expert trajectories in $\mathcal{D}$ and pseudo-expert trajectories in $\mathcal{D}_{\text{sim}}$.
 As for the vocabulary scoring-based planner, the learning objective has additional prediction of reward signals $r$ that distill the evaluation metrics, \eg. EPDMS as Eq.~\ref{eq:epdms}:
\begin{equation}
    \begin{aligned}
\label{eq:im_reward}
    \arg\min_{\theta}\mathbb{E}_{(a,o,r)\sim (\mathcal{D}\cup\mathcal{D}_{\text{sim}})}
    \Big[
    \lambda \mathcal{L}_{\text{im}}
    + \mathcal{L}_{\text{r}}\big(r,\, \pi_\theta(\hat{a}|o)\big)
    \Big], 
\end{aligned}
\end{equation}
here, $\mathcal{L}_{\text{r}}$ denotes reward loss, $\lambda $ is a weighting factor.

\begin{figure}[t!]
    \centering
    \includegraphics[width=1.\linewidth]{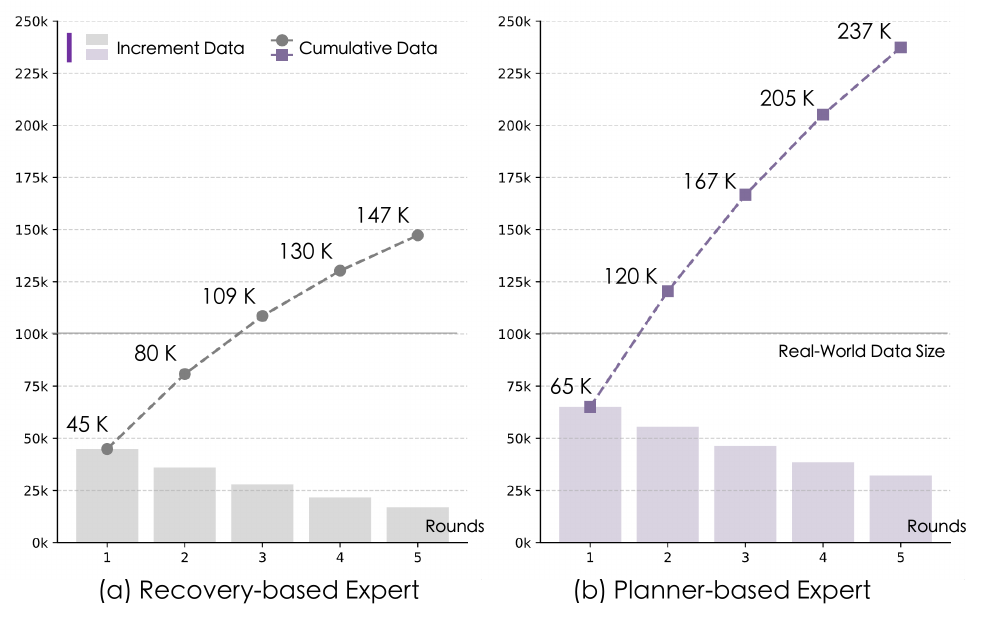}
    \caption
    {\textbf{Simulation data   % Status
    statistics across multiple sampling rounds.}
    \textbf{(a)} Recovery-based expert impose stronger constraints, leading to slower data accumulation than \textbf{(b)} Planner-based expert.
    }
    \label{fig:data}
    \vspace{-3pt}
\end{figure}

\boldparagraph{(2) Co-training with Rewards Only.}
For the vocabulary scoring-based planner, when the reward signal is well-aligned, the expert is theoretically unnecessary, as shown in Eq.~\ref{eq:im_reward}. The planner can explore directions to increase the reward without being restricted to a single expert trajectory. The co-training process can be formulated as:
\begin{equation} 
\arg\min_{\theta}\mathbb{E}_{(a,o,r)\sim \mathcal{D}} [\lambda \mathcal{L}_\text{im} + \mathcal{L}_r]
+ \mathbb{E}_{(o,r)\sim \mathcal{D}_{\text{sim}}} [\mathcal{L}_r].
\end{equation}
 We thus investigate purely reward-driven optimization for the scoring-based planner in simulation, to assess its potential for fully leveraging simulation data, and assess its potential for maximizing the utility of simulated data.

\section{Experiments}
\label{sec:exp}

\subsection{Setup and Protocols.}
\label{sec:exp_setup}

\boldparagraph{Real-world Datasets.} We use the \cmtt{navtrain} split from NAVSIM~\cite{Dauner2024navsim}, built upon the largest publicly available annotated driving dataset nuPlan~\cite{karnchanachari2024nuplan}, and comprises 100K interactive real-world scenarios.

\boldparagraph{Simulation Data Curation.} We construct simulated scenarios on \cmtt{navtrain} split. 3DGS models of blocks with $\mathrm{PSNR}<27$ in novel-view results are removed ensure reconstruction quality. 
For ego trajectory perturbation, we choose the plausible candidates from the vocabulary of 16,384 trajectories from~\cite{li2025gtrs}, with $\mathrm{EPDMS} \ge 0.8$ and relative heading $|\Delta \theta| \le 20^\circ$, where $\mathrm{EPDMS}$ is detailed in Eq.~\ref{eq:epdms}. During spatial sampling by trajectory endpoint, the longitudinal and lateral perturbation ranges are set as $r_\mathrm{lon}=20\mathrm{m},
r_\mathrm{lat}=2.0\mathrm{m}$, with step sizes $
\delta_\mathrm{lon}=5\mathrm{m},
\delta_\mathrm{lat}=0.5\mathrm{m}$. 
For recovery-based trajectory generation, we use all human expert trajectories in \cmtt{navtrain} split as the retrieval vocabulary. For planner-based trajectory generation, we select the SOTA rule-based planner in nuPlan~\cite{karnchanachari2024nuplan},  PDM-Closed~\cite{dauner2023pdmclosed} as the expert. 
Only pseudo-expert trajectories with decent sub-metrics of $\mathrm{EPDMS}$ are considered valid.
Due to computational limits, for each type of pseudo-expert, we perform up to five random samplings of valid trajectories with no overlaps
from \cmtt{navtrain} split for rendering. 
Through multiple samplings, expert trajectory demonstrations are progressively accumulated.
As shown in Fig.~\ref{fig:data}, the two pseudo-expert methods exhibit different success rates, with the planner-based expert achieving higher efficiency.
In total, we generate \textbf{147K} recovery-based simulation scenes and \textbf{237K} planner-based simulation scenes. For more details, please refer to Sec.~\ref{sec:supp_imple_data} in the supplementary
% ~\ref{sec:supp_imple_data}  {\color{cvprblue}B.1}

% \vspace{-0.1in}
\begin{table*}[htbp]
\centering
\caption{\textbf{Performance on the NAVSIM-v2 \cmtt{navhard} Leaderboard.} PDM-Closed uses ground-truth symbolic inputs for planning, while other methods rely on sensor data. ($*$: pseudo-expert supervision; $\dagger$: reward scoring; S.: per-stage EPDM score.)}
\vspace{-0.1in}
\scriptsize
\resizebox{1\textwidth}{!}{
\begin{tabular}{l|c|c|*{1}{p{0.03\textwidth}<{\centering}} |*{4}{p{0.03\textwidth}<{\centering}} |*{5}{p{0.03\textwidth}<{\centering}} |l|l}

    \toprule
    \textbf{Method}
    & \textbf{Backbone}
    & \textbf{Sim.}
    & \textbf{Stage}
    & \textbf{NC}↑
    & \textbf{DAC}↑
    & \textbf{DDC}↑
    & \textbf{TLC}↑
    & \textbf{EP}↑
    & \textbf{TTC}↑
    & \textbf{LK}↑
    & \textbf{HC}↑
    & \textbf{EC}↑
    & \textbf{S.}↑
    & \textbf{EPDMS}↑  \\
    \midrule
    
     \makecell[l]{PDM-Closed~\cite{dauner2023pdmclosed}}  & - & - &   \makecell{S 1 \\ S 2} & \makecell{94.4 \\ 88.1} & \makecell{98.8 \\ 90.6} & \makecell{100 \\ 96.3} & \makecell{99.5 \\ 98.5} & \makecell{100 \\ 100} & \makecell{93.5 \\ 83.1} & \makecell{99.3 \\ 73.7} & \makecell{87.7 \\ 91.5} & \makecell{36.0 \\ 25.4} &\makecell{- \\ -} & 51.3    \\
    \midrule
    \rowcolor[HTML]{EEE5F4}\hline
     \multicolumn{15}{c}{\textit{{Regeression-based Planner}}}\\
    \multirow{4}{*}{
    \makecell[l] {
    LTF~\cite{kashyap2022transfuser}}} & \multirow{4}{*}{ResNet34} & \textit{w/o} &   \makecell{S 1 \\ S 2}  &
    \makecell{97.3\\79.4} 
    & \makecell{80.2\\69.0} 
    & \makecell{97.8\\85.6}
    & \makecell{99.3\\98.5} 
    & \makecell{83.4\\83.8} 
    & \makecell{96.2\\76.7} 
    & \makecell{92.9\\47.9} 
    & \makecell{97.8\\97.0} 
    & \makecell{71.1\\70.6} 
    & \makecell[l]{61.3\\39.2}
    & 24.4    \\
    \cmidrule{3-15}
    
    &  & \cellcolor[rgb]{0.9,0.95,1}\textit{w/}{$^\ast$} 
    & \cellcolor[rgb]{0.9,0.95,1}\makecell{S 1 \\ S 2}  
    & \cellcolor[rgb]{0.9,0.95,1}\makecell{96.1 \\85.5}
    & \cellcolor[rgb]{0.9,0.95,1}\makecell{85.3 \\66.9}
    & \cellcolor[rgb]{0.9,0.95,1}\makecell{99.4 \\91.6} 
    & \cellcolor[rgb]{0.9,0.95,1}\makecell{99.3 \\99.1} 
    & \cellcolor[rgb]{0.9,0.95,1}\makecell{84.7 \\93.0} 
    & \cellcolor[rgb]{0.9,0.95,1}\makecell{94.7 \\81.1} 
    & \cellcolor[rgb]{0.9,0.95,1}\makecell{93.6 \\58.3} 
    & \cellcolor[rgb]{0.9,0.95,1}~\makecell{97.6 \\95.1}
    & \cellcolor[rgb]{0.9,0.95,1}\makecell{77.3 \\42.9}
    & \cellcolor[rgb]{0.9,0.95,1}\makecell[l]{\textbf{66.3}\textcolor[rgb]{0,0.3,0.6}{{\tiny +5.0}}  \\ \textbf{44.8}\textcolor[rgb]{0,0.3,0.6}{{\tiny +5.6}}  }
    & \cellcolor[rgb]{0.9,0.95,1}\textbf{30.2} \textcolor[rgb]{0,0.3,0.6}{{↑24\%}}  \\
        \rowcolor[HTML]{EEE5F4}\hline
     \multicolumn{15}{c}{\textit{{Diffusion-based Planner}\quad \quad}} \\
    \multirow{4}{*}{{\makecell[l]{
    DiffusionDrive~\cite{liao2025diffusiondrive}}} } & \multirow{4}{*}{ResNet34} & \textit{w/o} &   \makecell{S 1 \\ S 2}  
    & \makecell{96.8 \\ 80.1} 
    & \makecell{ 86.0\\72.8 } 
    & \makecell{ 98.8\\ 84.4}
    & \makecell{ 99.3\\98.4} 
    & \makecell{ 84.0\\85.9 } 
    & \makecell{ 95.8\\76.6 } 
    & \makecell{ 96.7\\46.4 } 
    & \makecell{ 97.6\\96.3 } 
    & \makecell{79.6\\72.8 } 
    & \makecell[l]{ 66.7\\40.5 }
    & \cellcolor[rgb]{1,1,1}27.5    \\
    \cmidrule{3-15}

    &  & \cellcolor[rgb]{0.9,0.95,1}\textit{w/}{$^\ast$} 
    & \cellcolor[rgb]{0.9,0.95,1}\makecell{S 1 \\ S 2}  
    & \cellcolor[rgb]{0.9,0.95,1}\makecell{97.4 \\86.4}
    & \cellcolor[rgb]{0.9,0.95,1}\makecell{88.7 \\72.1}
    & \cellcolor[rgb]{0.9,0.95,1}\makecell{99.3 \\92.9} 
    & \cellcolor[rgb]{0.9,0.95,1}\makecell{99.3 \\98.5} 
    & \cellcolor[rgb]{0.9,0.95,1}\makecell{82.8 \\92.1} 
    & \cellcolor[rgb]{0.9,0.95,1}\makecell{96.9 \\80.6} 
    & \cellcolor[rgb]{0.9,0.95,1}\makecell{98.0 \\60.8} 
    & \cellcolor[rgb]{0.9,0.95,1}~\makecell{97.3 \\95.4}
    & \cellcolor[rgb]{0.9,0.95,1}\makecell{59.6 \\31.9}
    & \cellcolor[rgb]{0.9,0.95,1}\makecell[l]{\textbf{67.5}\textcolor[rgb]{0,0.3,0.6}{{\tiny +0.8}} \\ \textbf{46.8}\textcolor[rgb]{0,0.3,0.6}{{\tiny +6.3}} }
    & \cellcolor[rgb]{0.9,0.95,1}\textbf{32.6} \textcolor[rgb]{0,0.3,0.6}{{↑19\%}}  \\
    \rowcolor[HTML]{EEE5F4}\hline
     \multicolumn{15}{c}{\textit{{Scoring-based Planner}\quad \quad\quad }} \\
    
    \multirow{8}{*}{{\makecell[l]{
    GTRS-Dense~\cite{li2025gtrs}}}} & 
    
    \multirow{4}{*}{ResNet34} & \textit{w/o} &   \makecell{S 1 \\ S 2}  
    &\makecell{ 99.3 \\ 92.8 } 
    & \makecell{ 96.6 \\ 88.6} 
    & \makecell{ 99.6 \\ 95.5 }
    & \makecell{ 100  \\ 99.4} 
    & \makecell{ 57.4 \\ 55.9} 
    & \makecell{ 99.5 \\ 91.3} 
    & \makecell{ 92.6 \\ 55.7} 
    & \makecell{ 89.5 \\ 91.1} 
    & \makecell{16.4 \\ 35.7} 
    & \makecell[l]{67.1\\55.8 }
    & 38.3    \\
    \cmidrule{3-15}

    &  & \cellcolor[rgb]{0.9,0.95,1}\textit{w/}${^\dagger}$ 
    & \cellcolor[rgb]{0.9,0.95,1}\makecell{S 1 \\ S 2} 
    & \cellcolor[rgb]{0.9,0.95,1}\makecell{97.6 \\94.3}
    & \cellcolor[rgb]{0.9,0.95,1}\makecell{96.4 \\92.7}
    & \cellcolor[rgb]{0.9,0.95,1}\makecell{99.3 \\95.1} 
    & \cellcolor[rgb]{0.9,0.95,1}\makecell{100 \\99.5} 
    & \cellcolor[rgb]{0.9,0.95,1}\makecell{75.7 \\80.2} 
    & \cellcolor[rgb]{0.9,0.95,1}\makecell{97.8 \\91.5} 
    & \cellcolor[rgb]{0.9,0.95,1}\makecell{93.3 \\56.2} 
    & \cellcolor[rgb]{0.9,0.95,1}~\makecell{97.3 \\90.6}
    & \cellcolor[rgb]{0.9,0.95,1}\makecell{32.9 \\28.3} 
    & \cellcolor[rgb]{0.9,0.95,1}\makecell[l]{ \textbf{72.4}\textcolor[rgb]{0,0.3,0.6}{{\tiny +5.3}} \\ \textbf{63.4}\textcolor[rgb]{0,0.3,0.6}{{\tiny +7.6}} }
    & \cellcolor[rgb]{0.9,0.95,1}\textbf{46.9} \textcolor[rgb]{0,0.3,0.6}{{↑22\%}} \\
    \cmidrule{2-15} &
    
    \multirow{4}{*}{V2-99}
    & \textit{w/o}&   \makecell{S 1 \\ S 2}  &
    \makecell{ 98.9 \\ 89.9} 
    & \makecell{ 94.9 \\ 90.5} 
    & \makecell{ 99.1\\ 94.1}
    & \makecell{ 100 \\ 99.3} 
    & \makecell{ 76.1\\ 77.6} 
    & \makecell{ 98.4\\ 88.5} 
    & \makecell{ 93.8 \\ 56.0} 
    & \makecell{ 94.9\\ 92.0} 
    & \makecell{ 37.8 \\ 30.2} 
    & \makecell[l]{ 70.4\\ 58.5}
    &   41.9    \\
    \cmidrule{3-15}

    & & \cellcolor[rgb]{0.9,0.95,1}\textit{w/}${^\dagger}$ 
    & \cellcolor[rgb]{0.9,0.95,1}\makecell{S 1 \\ S 2}  
    & \cellcolor[rgb]{0.9,0.95,1}\makecell{99.3 \\94.9}
    & \cellcolor[rgb]{0.9,0.95,1}\makecell{97.6 \\94.3}
    & \cellcolor[rgb]{0.9,0.95,1}\makecell{99.4 \\94.5} 
    & \cellcolor[rgb]{0.9,0.95,1}\makecell{100 \\99.3} 
    & \cellcolor[rgb]{0.9,0.95,1}\makecell{71.5 \\79.0} 
    & \cellcolor[rgb]{0.9,0.95,1}\makecell{99.6 \\92.6} 
    & \cellcolor[rgb]{0.9,0.95,1}\makecell{96.0 \\57.8} 
    & \cellcolor[rgb]{0.9,0.95,1}~\makecell{95.8 \\93.4}
    & \cellcolor[rgb]{0.9,0.95,1}\makecell{30.7 \\30.9}
    & \cellcolor[rgb]{0.9,0.95,1}\makecell[l]{\textbf{72.5}\textcolor[rgb]{0,0.3,0.6}{{\tiny +2.1}}\\ \textbf{65.4}\textcolor[rgb]{0,0.3,0.6}{{\tiny +6.9}} }
    & \cellcolor[rgb]{0.9,0.95,1}\textbf{48.0} \textcolor[rgb]{0,0.3,0.6}{↑15\%} \\
\bottomrule
\end{tabular}
}
\label{tab:navhard}
\end{table*}

\begin{table*}[htbp]
\centering
\caption{\textbf{Performance on the NAVSIM-v2 \cmtt{navtest} Leaderboard.} ($*$: pseudo-expert supervision; $\dagger$: reward scoring.)}
\vspace{-0.1in}
\scriptsize
\resizebox{\textwidth}{!}{
\begin{tabular}{l | c|c |*{4}{p{0.03\textwidth}<{\centering}}|*{5}{p{0.03\textwidth}<{\centering}}|l}

    \toprule
    \textbf{Method}
    & \textbf{Backbone}
    & \textbf{Sim.}
    & \textbf{NC}↑ 
    & \textbf{DAC}↑
    & \textbf{DDC}↑
    & \textbf{TLC}↑
    & \textbf{EP}↑
    & \textbf{TTC}↑
    & \textbf{LK}↑
    & \textbf{HC}↑
    & \textbf{EC}↑ 
    & \textbf{EPDMS}↑  \\
    \midrule
    
     \makecell[l]{Human Agent}  & - & -  & 100 & 100 & 99.8  & 100 & 87.4 & 100 & 100 & 98.1 & 90.1 & 90.3    \\
    \midrule
            \rowcolor[HTML]{EEE5F4}\hline
     \multicolumn{13}{c}{\textit{{Regression-based Planner}}} \\
    
    \multirowcell{2}[-0.5ex][l]{
    LTF~\cite{kashyap2022transfuser}} & \multirowcell{2}[-0.8ex][c]{ResNet34} 
    & \textit{w/o} 
    & 97.7
    & 94.0
    & 99.3
    & 99.8
    & 87.2 
    & 96.7 
    & 95.5 
    & {98.3} 
    & 82.9 
    & 81.5   \\
    \cmidrule{3-13}

     &  & \cellcolor[rgb]{0.9,0.95,1}\textit{w/}${^\ast}$
     & \cellcolor[rgb]{0.9,0.95,1}98.3
     & \cellcolor[rgb]{0.9,0.95,1}95.6
     & \cellcolor[rgb]{0.9,0.95,1}99.6
     & \cellcolor[rgb]{0.9,0.95,1}99.8
     & \cellcolor[rgb]{0.9,0.95,1}87.1 
     & \cellcolor[rgb]{0.9,0.95,1}97.5
     & \cellcolor[rgb]{0.9,0.95,1}97.2
     & \cellcolor[rgb]{0.9,0.95,1}98.3
     & \cellcolor[rgb]{0.9,0.95,1}88.2
     & \cellcolor[rgb]{0.9,0.95,1}\textbf{84.4} \textcolor[rgb]{0,0.3,0.6}{{\tiny +2.9}}  \\
            \rowcolor[HTML]{EEE5F4}\hline
     \multicolumn{13}{c}{\textit{{Diffusion-based Planner\quad \quad}}} \\

   \multirowcell{2}[-0.5ex][l]{
   DiffusionDrive~\cite{liao2025diffusiondrive}} & \multirowcell{2}[-0.8ex][c]{ResNet34} 
    & \textit{w/o} 
    & {98.4}
    & 95.5
    & 99.5 
    & 99.8 
    & 87.5 
    & 97.5 
    & 96.9
    &  {98.4}
    & {87.7} 
    & 84.2    \\
    \cmidrule{3-13}

     &  & \cellcolor[rgb]{0.9,0.95,1}\textit{w/}${^\ast}$
     & \cellcolor[rgb]{0.9,0.95,1}98.5
     & \cellcolor[rgb]{0.9,0.95,1}97.1
     & \cellcolor[rgb]{0.9,0.95,1}99.6
     & \cellcolor[rgb]{0.9,0.95,1}99.8
     & \cellcolor[rgb]{0.9,0.95,1}87.4 
     & \cellcolor[rgb]{0.9,0.95,1}97.8
     & \cellcolor[rgb]{0.9,0.95,1}98.0
     & \cellcolor[rgb]{0.9,0.95,1}98.3
     & \cellcolor[rgb]{0.9,0.95,1}87.5
     & \cellcolor[rgb]{0.9,0.95,1}\textbf{85.9} \textcolor[rgb]{0,0.3,0.6}{{\tiny +1.7}}    \\ 
    \rowcolor[HTML]{EEE5F4}\hline
     \multicolumn{13}{c}{\textit{{Scoring-based Planner\quad \quad\quad }}} \\

    \multirowcell{4}[-2.5ex][l]{
    GTRS-Dense~\cite{li2025gtrs}} & 
    \multirowcell{2}[-0.8ex][c]{ResNet34} 
    & \textit{w/o} 
    & {97.6} 
    & 97.5
    & 99.0
    & {99.9}
    & 87.9 
    & 97.0 
    & 95.9 
    & {97.5} 
    & 55.9 
    & 82.3    \\
    \cmidrule{3-13}
      &  & \cellcolor[rgb]{0.9,0.95,1}\textit{w/}${^\dagger}$
     & \cellcolor[rgb]{0.9,0.95,1}98.4
     & \cellcolor[rgb]{0.9,0.95,1}98.8
     & \cellcolor[rgb]{0.9,0.95,1}99.4
     & \cellcolor[rgb]{0.9,0.95,1}99.9
     & \cellcolor[rgb]{0.9,0.95,1}88.0
     & \cellcolor[rgb]{0.9,0.95,1}98.0
     & \cellcolor[rgb]{0.9,0.95,1}96.5
     & \cellcolor[rgb]{0.9,0.95,1}97.6
     & \cellcolor[rgb]{0.9,0.95,1}58.0
      & \cellcolor[rgb]{0.9,0.95,1}\textbf{84.6} \textcolor[rgb]{0,0.3,0.6}{{\tiny +2.3}}    \\
     \cmidrule{2-13} &
    \multirowcell{2}[-0.8ex][c]{V2-99} 
    & \textit{w/o} 
    & 97.6 
    & 98.5
    & {99.5}
    & {99.9} 
    & {89.5} 
    & 97.2 
    & {96.8} 
    & 97.2 
    & 57.2 
    & 84.0    \\
    \cmidrule{3-13}
      &  & \cellcolor[rgb]{0.9,0.95,1}\textit{w/}${^{\dagger}}$ 
     & \cellcolor[rgb]{0.9,0.95,1}97.6
     & \cellcolor[rgb]{0.9,0.95,1}98.9
     & \cellcolor[rgb]{0.9,0.95,1}99.6
     & \cellcolor[rgb]{0.9,0.95,1}99.9
     & \cellcolor[rgb]{0.9,0.95,1}89.9
     & \cellcolor[rgb]{0.9,0.95,1}97.1
     & \cellcolor[rgb]{0.9,0.95,1}97.2
     & \cellcolor[rgb]{0.9,0.95,1}97.7
     & \cellcolor[rgb]{0.9,0.95,1}58.7
      & \cellcolor[rgb]{0.9,0.95,1}\cellcolor[rgb]{0.9,0.95,1}\textbf{84.6} \textcolor[rgb]{0,0.3,0.6}{{\tiny +0.8}}    \\

\bottomrule
\end{tabular}
}
\vspace{-5pt}
\label{tab:navtest}
\end{table*}

\boldparagraph{Benchmark and Metrics.} We utilize two benchmarks in NAVSIMv2~\cite{Cao2025navsimv2} for end-to-end model evaluation, 
including \cmtt{navhard} and \cmtt{navtest}. \cmtt{navhard} is the official two-stage evaluation benchmark, which contains 244 challenging real-world scenarios in the first stage and corresponding 4,164 synthetic scenarios generated by 3DGS in the second stage. \cmtt{navtest} is a one-stage evaluation benchmark, containing a large number of 12,146 real-world scenarios. \cmtt{navhard} focuses on assessing the model’s closed-loop performance in safety-critical situations, while \cmtt{navtest} emphasizes generalization across diverse driving conditions. The two benchmarks share a rule-based planning metric, $\mathrm{EPDMS}$~\cite{li2025hydramdp++}, with several sub-metrics:
\begin{equation}
\mathrm{EPDMS} = 
\underbrace{\left(
\prod_{m \in \mathcal{M}_\text{pen}} S_m \right)
}_{\text{penalties}}
\cdot
\underbrace{\left(
\frac{ \sum_{m \in \mathcal{M}_\text{avg}} w_m  S_m }
     { \sum_{m \in \mathcal{M}_\text{avg}} w_m }\right)
}_{\text{weighted average}},
\label{eq:epdms}
\end{equation}
where $S_m$ is the sub-metric: penalty terms set $\mathcal{M}_\text{pen}$ includes No-at-fault Collisions (NC), Drivable Area Compliance (DAC), Driving Direction Compliance (DDC), and Traffic Light Compliance (TLC); weighted average terms set $\mathcal{M}_\text{avg}$ includes Time-to-Collision (TTC), Ego Progress (EP), Lane Keeping (LK), History Comfort (HC), and extended comfort (EC). Note that $\mathrm{EPDMS}$ in \cmtt{navhard} further incorporates several modifications, \eg, two-stage aggregation, reactive traffic simulation, and the exclusion of penalties in cases where the human expert driver also fails.

\boldparagraph{Models and Training.} To thoroughly validate our model-agnostic simulation data, we select one representative open-source model for each paradigm discussed in Sec.~\ref{sec:cotraining}: the regression-based LTF~\cite{kashyap2022transfuser}, the diffusion-based DiffusionDrive~\cite{liao2025diffusiondrive}, and the scoring-based GTRS-Dense~\cite{li2025gtrs}. We adopt their official implementations with two modifications: the input image resolution is unified to $2048\times512$, and the LiDAR inputs are removed to align with the \cmtt{navhard} evaluation setting. All models are trained from scratch on NVIDIA H20-3e GPUs. To ensure fair comparison, each model follows the same training strategy, whether using only the \cmtt{navtrain} split or augmented with simulation data. For
more details, please refer to Sec.~\ref{sec:supp_imple_model} in supplementary.
% ~\ref{sec:supp_imple_model} {\color{cvprblue}B.2}

\subsection{Leaderboard Results}
\label{sec:exp_leaderboard}
Tab.~\ref{tab:navhard} and Tab.~\ref{tab:navtest} present the leaderboard results of \name~sim-real co-training for the three planner paradigms on \cmtt{navhard} and \cmtt{navtest}, respectively.

\boldparagraph{Navhard Leaderboard.}
All models exhibit significant improvements in both Stage~1 and Stage~2. Notably, GTRS-Dense (V2-99) achieves a score of \textbf{48.0}, establishing a new SOTA on \cmtt{navhard}.
These results demonstrate that incorporating simulation data with an extended distribution substantially enhances model robustness in challenging and unseen environments. Notably, weaker baseline models, \ie, LTF and DiffusionDrive, benefit the most, with gains exceeding 20\%, indicating that sim-real co-training with simulation data effectively enables models to exploit dataset knowledge better and unlock their latent learning potential.

\begin{figure*}[t!]
    \centering
        \includegraphics[width=1\linewidth]{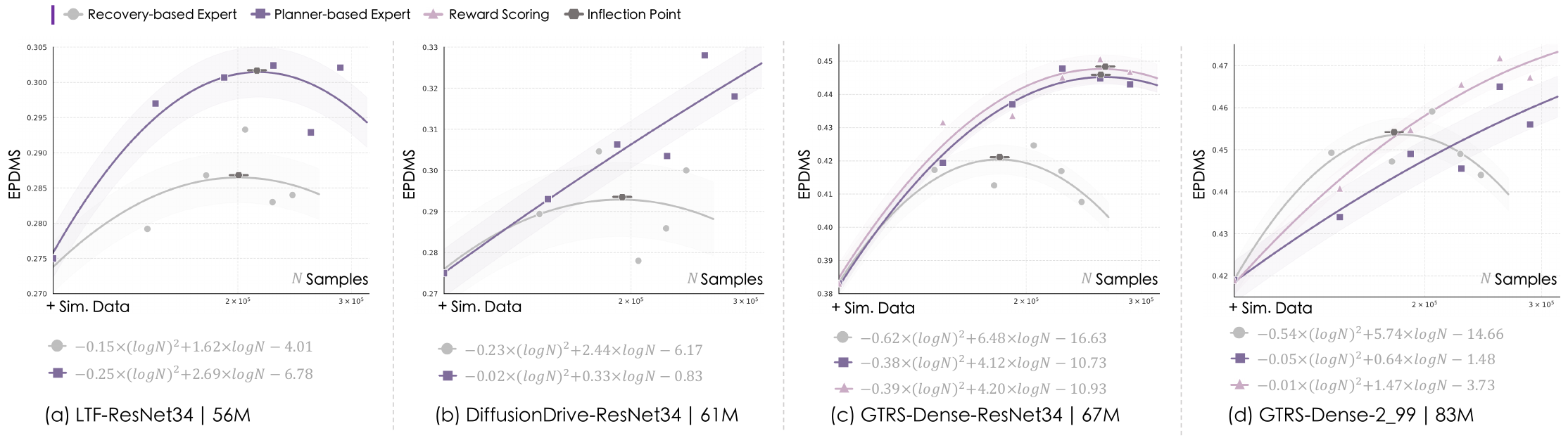}
    \vspace{-15pt}
    \caption{\textbf{Scaling dynamics across different planners and pseudo-expert trajectories.} 
    We visualize how simulation data scale and supervision signals influence the driving performance of various planners, where the infection point indicates learning plateau.
    }
    \label{fig:scaling}
    \vspace{-3pt}
\end{figure*}
\begin{figure*}[t!]
    \centering
    \captionsetup{type=figure}\includegraphics[width=\textwidth]{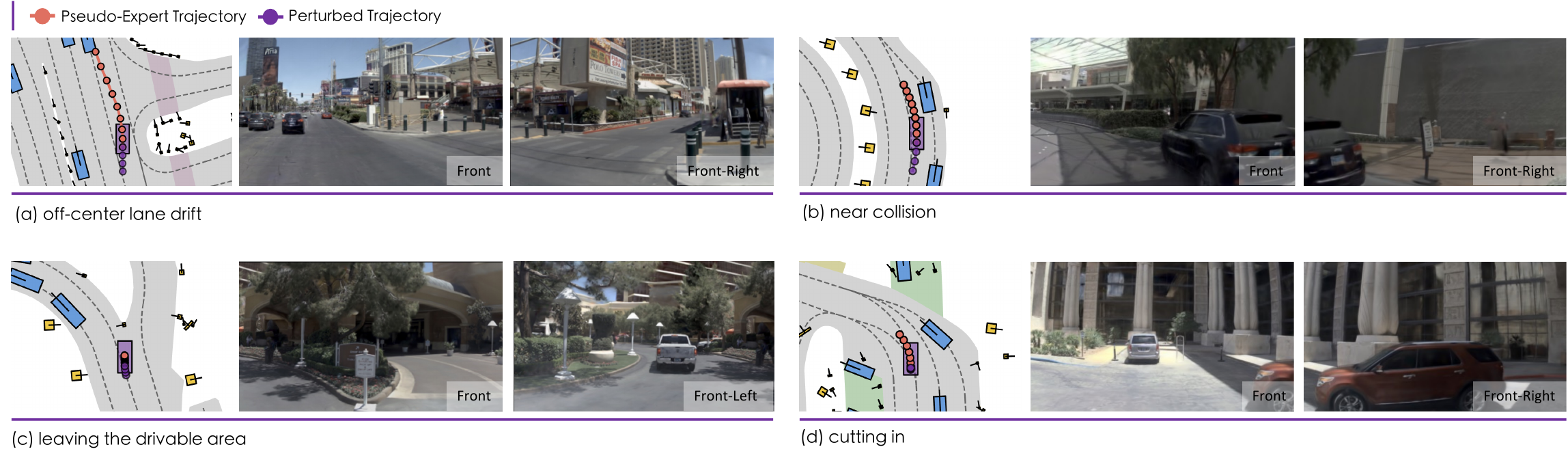} 
    \vspace{-15pt}
    \caption{\textbf{Qualitative results of the simulation scenes on \cmtt{navtrain}}. Four representative simulation scenarios are shown, each mirroring a typical real-world OOD scene, with synthetic front-view and auxiliary key-view images provided.
    }
    \label{fig:scene}
    \vspace{-10pt}
\end{figure*}

\boldparagraph{Navtest Leaderboard.}
All models show consistent improvements of up to +2.9 points, demonstrating stronger performance under large-scale and diverse conditions. 
The quantitative results above highlight that our simulation data is \textbf{model-agnostic}, and that our general sim-real co-training achieves a \textbf{synergistic optimization of robustness and generalization}, which is essential for reliable closed-loop deployment in the real world~\cite{chen2023e2esurvey,liu2025reinforced}.

\subsection{Ablation and Data Scaling Analysis}

\boldparagraph{Data Scaling Curves of Different Planners.}  Due to the lack of prior work studying the scaling behavior of simulation data under a fixed amount of real-world data, we model the relationship between performance and the total data size (simulation + real) using a log-quadratic function:
\vspace{-1pt}
\begin{equation}
S(N) = a \log^2(N) + b \log(N) + c,
\end{equation}
where \(S(N)\) denotes the planner performance with total data size \(N\), and \(a\), \(b\), \(c\) are parameters fitted via nonlinear least squares:
\vspace{-4pt}
\begin{equation}
(a^*, b^*, c^*) = \arg\min_{a,b,c} \sum_{i=1}^{M} \big( S_i - S(N_i; a,b,c) \big)^2,
\end{equation}
where \(S_i\) is the observed performance at total data size \(N_i\), and \(M\) is the number of data points. If the data scaling trend exists, the quadratic coefficient \(a\) approaches zero, degenerating the model to a linear log relation; otherwise, the curve exhibits a parabolic shape with a visible saturation point. We evaluate four planners under 2 pseudo-experts following simulation data scaling in Fig.~\ref{fig:data} and select $\mathrm{EPDMS}$ in \cmtt{navhard} as \(S\).  The scaling curves and fitted log-quadratic functions are shown in Fig.~\ref{fig:scaling}, with the residual standard deviation depicted as an error band. Additionally, we conduct extra experiments using reward scoring only in planner-based simulation data for GTRS-Dense, proposed in Sec.~\ref{sec:cotraining}. Some classical trends can be observed in Fig.~\ref{fig:scaling}. For instance, comparing planner-based and reward-scoring settings in Fig.~\ref{fig:scaling} (c) and (d), larger models exhibit more pronounced data scaling trends under the same amount of data. Additional interesting and meaningful observations are highlighted below.

\boldparagraph{Pseudo-Expert Should Be Exploratory.}
For all planners, the scaling curves under recovery-based setting converge earlier and achieve lower performance compared to planner-based. The recovery-based expert always steers to the human log, limiting diversity as the simulation data grows from the same real scenarios. In contrast, the planner-based expert explores a broader range of possibilities and even produces feasible solutions in challenging situations. Consequently, the recovery-based expert only exhibits advantages in small-data regimes (Fig.~\ref{fig:scaling} (d)), likely because its trajectory distribution is human-like and better aligned with real-world data, making it easier to learn. In most cases, data scaling yields diminishing returns for recovery-based methods compared to planner-based ones, and performance may even degrade as visual artifacts accumulate and begin to dominate. These observations highlight the importance of exploratory behavior for pseudo-experts, which enhances the value of simulation data under scaling.

\boldparagraph{Multi-modality Modeling Sparks the Scaling.} Although the regression-based LTF and diffusion-based DiffusionDrive have comparable model sizes ($56\text{M}$ vs $61\text{M}$), they exhibit markedly different scaling properties for the planner-based in Fig.~\ref{fig:scaling} (a) and (b). 
For LTF, performance saturates and starts to degrade when the simulation-to-real ratio reaches 1:1, whereas DiffusionDrive exhibits an approximately linear improvement. This is due to the gradually increasing diversity of demonstrations from the same real scenario, which introduces an effectively multi-modal supervision problem. Single-mode regression struggles to model multi-peak distributions, leading to mode confusion and performance degradation, while diffusion models can capture multimodality, making them amenable to optimization under diverse supervision~\cite{cheng2025diffusionpolicy}. Since real-world autonomous driving is inherently a multi-peak problem~\cite{chen2024vadv2}, our simulation-scaling results underscore the importance of multimodal modeling for scalable real-world end-to-end autonomous driving.
\begin{table}[t]
\centering
\caption{\textbf{The effect of expert with simulated reward scoring} on \cmtt{navhard} using GTRS-Dense. ($\textbf{S}_{1/2}$:per-stage EPDM scores.)}
\small
\setlength{\tabcolsep}{2pt}
\begin{tabular}{c|cc|cc|ccc}
\toprule
\textbf{Backbone} 
& \textbf{Real} 
& \textbf{Expert} 
& \textbf{Sim.} 
& \textbf{Expert} 
& $\textbf{S}_1 \uparrow$
& $\textbf{S}_2 \uparrow$
& $\textbf{EPDMS} \uparrow$ 
\\
\midrule
\multirow{4}{*}{ResNet34} & \cellcolor[rgb]{0.95,0.95,0.95}\checkmark  & \cellcolor[rgb]{0.95,0.95,0.95}\checkmark  & & 
&67.1 & 55.8 &38.3  \\ 
& \cellcolor[rgb]{0.95,0.95,0.95}\checkmark   & \cellcolor[rgb]{0.95,0.95,0.95}  & & 
&66.8 & 54.9  & 37.6 \\
\cmidrule{2-8}
& \checkmark & \checkmark  & \cellcolor[rgb]{0.95,0.95,0.95}\checkmark   &\cellcolor[rgb]{0.95,0.95,0.95}\checkmark
 &
70.8&\textbf{63.9}&46.1 \\
& \checkmark  & \checkmark  & \cellcolor[rgb]{0.95,0.95,0.95}\checkmark   &\cellcolor[rgb]{0.95,0.95,0.95}
 &\textbf{72.4}&63.4&\textbf{45.9}\\
\midrule
\multirow{4}{*}{V2-99} & \cellcolor[rgb]{0.95,0.95,0.95}\checkmark  & \cellcolor[rgb]{0.95,0.95,0.95}\checkmark   & & 
&70.4 & 58.5 & 41.9  \\
& \cellcolor[rgb]{0.95,0.95,0.95}\checkmark   & \cellcolor[rgb]{0.95,0.95,0.95} &   &  
 &68.1&55.6&38.8\\
\cmidrule{2-8}
& \checkmark & \checkmark  & \cellcolor[rgb]{0.95,0.95,0.95}\checkmark   &\cellcolor[rgb]{0.95,0.95,0.95}\checkmark 
 &71.7&\textbf{65.5}&47.7\\
& \checkmark  & \checkmark  & \cellcolor[rgb]{0.95,0.95,0.95}\checkmark   &\cellcolor[rgb]{0.95,0.95,0.95} 
 &\textbf{72.5}&65.4&\textbf{48.0}\\
\bottomrule
\end{tabular}
\vspace{-5pt}
\label{tab:imi}
\end{table}

\boldparagraph{Reward is All You Need.} In Fig.~\ref{fig:scaling} (c) and (d), for the scoring-based GTRS-Dense planner, 
reward signals alone, without expert trajectories in simulation, yields even better performance. To further analyze this, we conduct reward-only training on real-world data only,
which instead leads to performance degradation, as shown in Tab.~\ref{tab:imi}. These results indicate that with sufficient expert supervision to stabilize the optimization direction, the reward guidance is better. The model benefits from the feedback from rewards during its exploration and interaction within the environment~\cite{cusumano2025selfplay}.

\boldparagraph{Effect of Reactive Simulation.}
\begin{table}[t]
\centering
\small
\caption{\textbf{The effect between non-reactive \vs reactive data simulation} on \cmtt{navhard} using GTRS-Dense, across sampling rounds.}
\setlength{\tabcolsep}{5pt}

\begin{tabular}{l|cr|cc}
\toprule
\multirowcell{2}[0ex][c]{\textbf{Type}}& 
\multirowcell{2}[0ex][c]{\textbf{\#Round}}&  
\multirowcell{2}[0ex][l]{\textbf{\#Sim.}}&
\multicolumn{2}{c}{\textbf{EPDMS}↑} \\ 
& & & {\footnotesize\textbf{ResNet34}} & {\footnotesize\textbf{V2-99}} \\
\midrule

Non-Reactive      
& 2
& 141K
& 43.7
& 45.6\\
Reactive  
& 2
& 120K
& \textbf{44.4}
& \textbf{46.7} \\
\midrule
\textcolor[rgb]{0.5,0.5,0.5}{Reactive}
& \textcolor[rgb]{0.5,0.5,0.5}{3}
& \textcolor[rgb]{0.5,0.5,0.5}{167K}
& \textcolor[rgb]{0.5,0.5,0.5}{45.0}
& \textcolor[rgb]{0.5,0.5,0.5}{47.9} \\
\bottomrule
\end{tabular}
\vspace{-5pt}
\label{tab:react}
\end{table}
To isolate the effect of reactive traffic, we compare reward-scoring GTRS-Dense on \cmtt{navhard} using non-reactive vs. reactive simulation data (Tab.~\ref{tab:react}). Two rounds of non-reactive sampling yield 141K trajectories, \ie, 21K more valid samples due to fewer collision rates, yet provide no performance improvement in $\mathrm{EPDMS}$. When reactive simulation reaches the third round, it produces 167K samples yet delivers consistent and significant $\mathrm{EPDMS}$ gains across both model sizes. These results indicate that reactive agent dynamics enhance the realism and diversity of traffic interactions, thereby increasing the effectiveness of simulation data. Detailed scaling curves are shown in Sec.~\ref{sec:supp_results_react} in the supplementary material.
% Sec.~\ref{sec:supp_results_react} {\color{cvprblue}C.3}

\subsection{Qualitative Results of the Simulation Scenes} 

We present qualitative visualizations from the simulation data in Fig.~\ref{fig:scene}, showcasing four representative OOD scenarios used to train the policy. 
These scenarios mirror typical real-world driving challenges where learned policies tend to struggle, including 
(a) off-center lane drift, 
(b) near collision, 
(c) departure, 
and (d) cutting in cases.
Each scenario is illustrated with a top-down view showing the pseudo-expert trajectory as supervision, and the deviating Perturbed Trajectory as history actions, alongside the synthetic front-view as sensory input to the policy. 
For instance, scenario (b) requires the policy to adaptively avoid the collision in short horizon. See supplementary Sec.~\ref{sec:supp_results_qual} for extended results.
% ~\ref{sec:supp_results_qual} {\color{cvprblue}C.2}

\section{Conclusion}

In this paper, we introduce \name, a sim–real learning system that reveals how scalable simulation can amplify the value of real-world datasets for end-to-end autonomy.
For the simulation data pipeline, we first generate pseudo-expert demonstrations from potential OOD states by ego perturbation within reactive environments. Toward the real-world simulation, the associated high-fidelity multi-view observations are rendered with 3DGS engine. 
Upon the simulated data, sim-real co-training produces synergistic improvements in robustness and generalization for various planners on challenging real-world benchmarks, up to +8.6 $\mathrm{EPDMS}$ on \cmtt{navhard} and +2.9 \cmtt{navtest}. 
Remarkably, the sim–real system scales clearly and predictably with increased simulation while keeping the real-world corpus fixed. 
We further uncover that exploration and interaction contribute to more effective simulation, and that multi-modal planners strengthen their scaling behavior.
We hope \name~will inspire the community to further explore real-world simulation for data scaling.

We provide additional experiments in the supplementary material, such as multi-expert ensemble and scaling with varying real data. It also provides extended discussions on related work, limitations, and broader impact.

\section*{Acknowledgments}
This study is supported by the National Key R\&D Program of China (2022ZD0117901), the National Natural Science Foundation of China (Grant No. 62373355, 62236010), and the Beijing Natural Science Foundation (Grant L252033). This work is in part supported by the JC STEM Lab of Autonomous Intelligent Systems funded by The Hong Kong Jockey Club Charities Trust. We also appreciate the generous research sponsor from Xiaomi.

We extend our gratitude to Naiyan Wang, Yunsong Zhou, Hanxue Zhang, and the rest of the members from OpenDriveLab and Xiaomi Embodied Intelligence Team for their profound support.

{
    \small
    \bibliographystyle{ieeenat_fullname}
    \bibliography{bib_short,main}
}

\clearpage
\newpage
\appendix

\clearpage
\setcounter{page}{1}
\maketitlesupplementary

\titlecontents{section}
  [1.5em]
  {\vspace{1ex}}
  {\bfseries\contentslabel{1.7em}}
  {}
  {\hfill\textbf\contentspage}

\titlecontents{subsection}
  [4.0em]
  {\vspace{0.1ex}}
  {\contentslabel{2.4em}}
  {}
  {\titlerule*[0.5pc]{.}\contentspage}

\startcontents
{
\hypersetup{linkcolor=black}
\printcontents{}{1}{}
\noindent\hrulefill
}
\vspace{5pt}

We outline the supplementary material as follows.
In Sec.~\ref{sec:supp_related}, we first provide additional discussions of related work.
Sec.~\ref{sec:supp_imple} presents further implementation details regarding data curation and models.
In Sec.~\ref{sec:supp_results_ext}, we report \textbf{extended experimental analyses and qualitative results} corresponding to the main experiments 
(Sec.~\ref{sec:exp}). 
In Sec.~\ref{sec:supp_results_add}, we further present \textbf{additional experimental studies}. 
Finally, Sec.~\ref{sec:supp_limit} discusses the limitations and broader impacts, and Sec.~\ref{sec:supp_license} lists the licenses of all utilized assets. 
% ~\ref{sec:exp} {\color{cvprblue}3}

\section{Related Work} \label{sec:supp_related}

\subsection{End-to-End Autonomous Driving}

End-to-end system maps directly from raw sensor inputs to planning~\cite{chen2023e2esurvey}.
Early works adopt regression-based planning and gradually shift from extra task branches~\cite{kashyap2022transfuser, wu2022tcp,jia2023thinktwice} to unifying perception, prediction, and planning under joint supervision~\cite{hu2023uniad, jiang2023vad, weng2024paradrive, liu2025hybrid, jia2025drivetransformer}.
Recent work has moved toward generative approaches. Diffusion-based systems~\cite{liao2025diffusiondrive,xing2025goalflow,li2025recogdrive,li2025trajhF,dang2026drivefine} are framed as a conditional denoising process, enabling diverse and high-fidelity trajectories.
Concurrently, trajectory scoring has emerged as an efficient alternative, ranking candidate trajectories under spline curves~\cite{casas2021mp3,hu2022st}, discretized tokens~\cite{chen2024vadv2,gao2025rad}, clustered human trajectories~\cite{li2025hydranext,li2025gtrs},  or predicted proposals~\cite{sun2025sparsedrive,guo2025ipad, li2025wote,xie2026latentvla}.
Moreover, it offers a natural interface for reward- or cost-based optimization~\cite{li2025recogdrive, liu2025reinforced, li2025evadrive, li2025ztrs, gao2025rad, gao2026takevla, gao2026perlad}, enabling reinforced improvements. Still, supervision on logged end-to-end data limits training to the expert’s open-loop distribution, causing compromised learning in the drifted state of sensory data. We address this by introducing a scalable simulation framework that reactively generates end-to-end pairs using pseudo-experts or rewards, allowing end-to-end systems across any above paradigms to bootstrap extra supervision from existing training data.

\subsection{Scene Simulation for Driving}

Scene simulation, including traffic behavior simulation and sensor simulation, has long been a key topic in autonomous driving research. 
For traffic simulation, existing works~\cite{Cao2025navsimv2, gao2025rad, zhou2024hugsim} leverage rule-based planners like IDM~\cite{treiber2000idm}, or diffusion-based generators~\cite{jiang2024scenediffuser,zhou2025nexus} to simulate plausible interaction of traffic agents. 
For sensor simulation, traditional graphics-based simulators~\cite{dosovitskiy17carla,li2022metadrive} suffer from a significant sim-to-real gap, which limits the real-world deployment of trained planners. Recent data-driven efforts follow two directions: some approaches~\cite{zhou2025safemvdrive, xu2025challenger,alhaija2025cosmos-transfer1,you2024bench2drive-r, wang2024drivedreamer,li2024drivingdiffusion, zhou2024simgen} attempt to generate sensor data in unseen scenarios via video generation models conditioned on 3D bounding boxes, HD maps or BEV maps. 
Other works focus on scene reconstruction~\cite{ljungbergh2024neuroncap, yan2024streetgs, chen2025omnire, zhou2024hugsim, jiang2025realengine, Cao2025navsimv2,xie2024vid2sim} to build photorealistic simulators for novel-view synthesis, using techniques like Neural Radiance Fields (NeRF) and 3D Gaussian Splatting (3DGS). 
Although these methods achieve impressive visual results, most are primarily designed for closed-loop evaluation or visual augmentation.
Our work aims to explore how to use traffic and sensor simulation to generate realistic scenes with feasible demonstrations for planner training, and its impacts on planner performances.
 
\subsection{Data Scaling for Driving}

Recent data scaling laws have driven major advances in foundation models \cite{kaplan2020scalinglaw,radford2021clip,zhai2022scalingvit,zhang2025mme}, but their impact on autonomous driving planning remains underexplored.
Prior researches~\cite{hwang2024emma,naumann2025e2escaling,zheng2024preliminaryscaling} demonstrate that increasing real-world data from thousands to millions of driving logs improves end-to-end planner performance, though improvements diminish. Dense supervision
from video predictions improves this data scaling efficiency \cite{li2025drivevla}. 
For planners operating on abstract BEV representations, industry efforts \cite{baniodeh2025motionscaling, huang2024drivegpt} demonstrate clear benefits from scaling real-world data and model capacity, while self-play \cite{cusumano2025selfplay,zhang2024asyself-play,cornelisse2025simagent} scales reinforcement learning via massive simulation to achieve strong zero-shot sim-to-real transfer.
However, most existing approaches rely on costly real-world collection or traffic simulation, which expands only in abstract state spaces rather than raw sensory domains. 3D rasterization provides a lightweight remedy, but suffers from information loss~\cite{feng2025rap}.
In contrast, we investigate scaling properties of planning directly through large-scale 3DGS sensory simulation, which bridges abstract traffic simulation and real-world perception, offering a scalable and realistic alternative to real-world data collection.

\subsection{Comparison with Online RL}

\begin{figure}[H]
    \vspace{-5pt}
    \centering
    \includegraphics[width=1\linewidth]{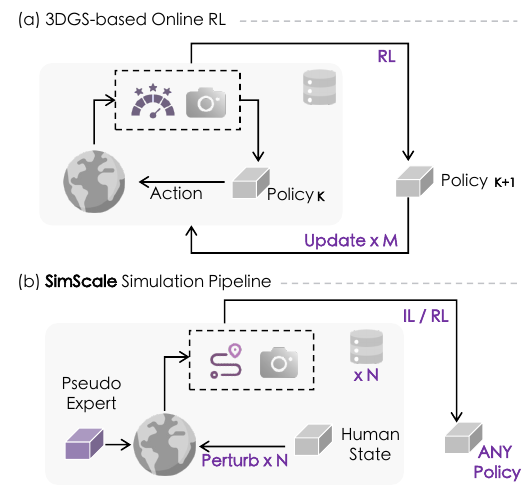}
    \caption
    {\textbf{Learning paradigm comparison of e2e autonomous driving between 3DGS-based Online RL and \name.}
    }
    \label{fig:position}
\end{figure}

Online reinforcement learning (RL)~\cite{jaeger2025carl, li2024think2drive, li2026plannerrft} in traffic simulators learn through exploration and feedback. Sensor simulation serves as a bridge to real-world environments, enabling real-world RL for autonomous driving, \ie, 3DGS-based Online RL~\cite{gao2025rad}, as shown in Fig.~\ref{fig:position} (a). In contrast, the proposed \name~introduces an alternative framework, as illustrated in Fig.~\ref{fig:position} (b), that generates OOD states with expert demonstrations at scale, enabling real-world simulations to support both imitation learning (IL) and reward learning (\ie Offline RL) across arbitrary planners.

\section{Extended Implementation Details} 
\label{sec:supp_imple}

\subsection{Simulation Data Curation}
\label{sec:supp_imple_data}

\begin{table}[t]
\centering
\caption{\textbf{Simulation Data Curation Pipeline Configurations.}}
\small
\resizebox{0.48\textwidth}{!}{

\begin{tabular}{lll}

\toprule
\multirowcell{1}[0ex][l]{\textbf{Term}}
& \multirowcell{1}[0ex][l]{\textbf{Notation}}
& \multirowcell{1}[0ex][l]{\textbf{Value}}
\\
\midrule

\multicolumn{3}{l}{\textbf{3DGS Data Engine}} \\

Peak signal-to-noise ratio of GS

& $\mathrm{PSNR}$
& $<27$\\

\midrule
\multicolumn{3}{l}{\textbf{Trajectory Perturbation}} \\
Clustered human traj vocabulary
& $\mathcal{V}_{\text{c}}$
& $16,384$
\\

Relative heading range to log 
& $\Delta\theta$
& $\pm 20^\circ$
\\

Longitudinal shift range 
& $r_\mathrm{lon}$
& $\pm 20\mathrm{m}$
\\

Lateral shift range
& $r_\mathrm{lat}$
& $\pm 2\mathrm{m}$
\\

Longitudinal sampling step
& $\delta_\mathrm{lon}$
& $5\mathrm{m}$
\\

Lateral sampling step 
& $\delta_\mathrm{lat}$
& $0.5\mathrm{m}$
\\

Reward filterer
& $\mathrm{R^{per}_{EPDMS}}$
& $ \ge 0.8$
\\

\midrule
\multicolumn{3}{l}{\textbf{Pseudo-Expert Trajectory Generation}} \\
Reward filterer
& $\mathrm{R^{exp}_{EPDMS}}$
& $\substack{ S_m=1, \forall m \neq \mathrm{EP} \\ S_{\mathrm{EP}} > 0.5 \quad~~~~ }$
\\

\multicolumn{3}{l}{\textbf{Recovery-based}} \\
Whole human traj vocabulary
& $\mathcal{V}_{\text{h}}$
& $103,288$
\\

\multicolumn{3}{l}{\textbf{Planner-based}} \\

Privileged planner
& $\mathbf{P}$
& PDM-Closed~\cite{dauner2023pdmclosed}

\\

\bottomrule
\end{tabular}
}
\label{tab:supp_data}
\end{table}

% Sec.~\ref{sec:exp_setup} {\color{cvprblue}3.1}
Tab.~\ref{tab:supp_data} summarizes the detailed terms with notation and value used in the simulation data curation process to supplement Sec.~\ref{sec:exp_setup} curation pipeline in the main paper. 
For pseudo-expert trajectory generation, we discard infeasible candidates to ensure valid supervision.
Specifically, all sub-metrics of $\mathrm{EPDMS}$ must be satisfied, with $S_\mathrm{EP}$ relaxed: $\mathrm{EP}\ge 0.5$, preventing biased driving styles. Qualitative results during curation are shown in Sec.~\ref{sec:supp_results_qual}

\subsection{Models and Training}
\label{sec:supp_imple_model}

\begin{table}[t]
\centering
\caption{\textbf{Model and Training Hyperparameters.}}
\small

\resizebox{0.48\textwidth}{!}{

\begin{tabular}{l|ccc}

\toprule
\multirowcell{1}[0ex][c]{\textbf{Hyperpara.}}

& \multirowcell{1}[0ex][c]{\textbf{LTF}~\cite{kashyap2022transfuser}}
& \multirowcell{1}[0ex][c]{\textbf{DiffusionDrive}~\cite{liao2025diffusiondrive}}
& \multirowcell{1}[0ex][c]{\textbf{GTRS-Dense}~\cite{li2025gtrs}}\\

\midrule

\multicolumn{4}{l}{\textbf{Model Configuration}} \\

Sensors

&$3 \times \text{Cam.}$
&$3 \times \text{Cam.}$
&$3 \times \text{Cam.}$

\\

Resolution

&$2048 \times 512$
&$2048 \times 512$
&$2048 \times 512$
\\

Horizon
&$4 \mathrm s$
&$4 \mathrm s$
&$4 \mathrm s$
\\

Frequency
&$2 \mathrm{Hz}$
&$2 \mathrm{Hz}$
&$10 \mathrm{Hz}$
\\

Backbone
& R34~\cite{he2016deep}
& R34~\cite{he2016deep}
& R34~\cite{he2016deep} / V99~\cite{lee2019centermask}
\\

Parameters
& $56\mathrm M$
& $61\mathrm M$
& $67\mathrm M$ / $83\mathrm M$
\\

Aux. Tasks
& Det. Seg.
& Det. Seg.
& None
\\

\midrule
\multicolumn{4}{l}{\textbf{Training Configuration}} \\
GPUs
& $8 \times \text{H20}$ 
& $8 \times \text{H20}$ 
& $32\times \text{H20}$ 
\\

Epochs
& $100$
& $100$
& $50$
\\

Total BS
& $512$
& $512$
& $352$
\\

Initial LR
& $2.83\times{10}^{-4}$
& $6\times{10}^{-4}$
& $4\times{10}^{-4}$
\\

Schedule
& Constant
& Cosine Decay
& Cosine Decay
\\

Optimizer
& Adam~\cite{kingma2014adam}
& AdamW~\cite{loshchilov2017adamw}
& AdamW~\cite{loshchilov2017adamw}
\\

\bottomrule
\end{tabular}
}

\label{tab:supp_imple}
\end{table}

\begin{table*}[htbp]
\centering
\caption{\textbf{Detailed Results on \cmtt{navhard}.} PDM-Closed uses ground-truth symbolic inputs for planning, while other methods rely on sensor data. (\textit{recovery / planner}: recovery-based / planner-based expert; \textit{reward}: reward-only scoring; S.: per-stage EPDM score.)}
\vspace{-0.1in}

\scriptsize
\resizebox{1\textwidth}{!}{

\begin{tabular}{l|l|c|*{1}{p{0.03\textwidth}<{\centering}} |*{4}{p{0.03\textwidth}<{\centering}} |*{5}{p{0.03\textwidth}<{\centering}} |c|c}

    \toprule
    \textbf{Method}
    & \textbf{Backbone}
    & \textbf{Sim.}
    & \textbf{Stage}
    & \textbf{NC}↑
    & \textbf{DAC}↑
    & \textbf{DDC}↑
    & \textbf{TLC}↑
    & \textbf{EP}↑
    & \textbf{TTC}↑
    & \textbf{LK}↑
    & \textbf{HC}↑
    & \textbf{EC}↑
    & \textbf{S.}↑
    & \textbf{EPDMS}↑  \\
    \midrule
    
     \makecell[l]{PDM-Closed~\cite{dauner2023pdmclosed}}  & - & - &   \makecell{S 1 \\ S 2} & \makecell{94.4 \\ 88.1} & \makecell{98.8 \\ 90.6} & \makecell{100 \\ 96.3} & \makecell{99.5 \\ 98.5} & \makecell{100 \\ 100} & \makecell{93.5 \\ 83.1} & \makecell{99.3 \\ 73.7} & \makecell{87.7 \\ 91.5} & \makecell{36.0 \\ 25.4} &\makecell{- \\ -} & 51.3    \\
    \midrule
    % \midrule
    \rowcolor[HTML]{EEE5F4}\hline
     \multicolumn{15}{c}{\textit{{Regeression-based Planner}}} \\
     
    \multirow{6}{*}{\makecell[l] {LTF~\cite{kashyap2022transfuser}}} 
    & \multirow{6}{*}{ResNet34} 
    & \textit{w/o} 
    & \makecell{S 1 \\ S 2}  
    & \makecell{97.3\\79.4} 
    & \makecell{80.2\\69.0} 
    & \makecell{97.8\\85.6}
    & \makecell{99.3\\98.5} 
    & \makecell{83.4\\83.8} 
    & \makecell{96.2\\76.7} 
    & \makecell{92.9\\47.9} 
    & \makecell{97.8\\97.0} 
    & \makecell{71.1\\70.6} 
    & \makecell[l]{61.3\\39.2}
    & 24.4    \\
    \cmidrule{3-15}

    & 
    & \textit{recovery}  
    & \makecell{S 1 \\ S 2}  
    & \makecell{ 96.4\\88.9 } 
    & \makecell{ 78.4\\71.1 } 
    & \makecell{ 98.9\\91.8 } 
    & \makecell{ 99.8\\99.0 }  
    & \makecell{ 80.5\\77.3 } 
    & \makecell{ 96.2\\85.5 } 
    & \makecell{ 92.7\\53.8 } 
    & \makecell{ 97.6\\96.8 } 
    & \makecell{ 78.2\\47.5 } 
    & \makecell{ 60.7\\\textbf{46.3} } 
    & 29.8   \\
    \cmidrule{3-15}
 
    &  
    & \cellcolor[rgb]{0.9,0.95,1}\textit{\textbf{planner}}
    & \cellcolor[rgb]{0.9,0.95,1}\makecell{S 1 \\ S 2}  
    & \cellcolor[rgb]{0.9,0.95,1}\makecell{ 96.1\\85.5 } 
    & \cellcolor[rgb]{0.9,0.95,1}\makecell{ 85.3\\66.9 } 
    & \cellcolor[rgb]{0.9,0.95,1}\makecell{ 99.4\\91.6 } 
    & \cellcolor[rgb]{0.9,0.95,1}\makecell{ 99.3\\99.1 }  
    & \cellcolor[rgb]{0.9,0.95,1}\makecell{ 84.7\\93.0 } 
    & \cellcolor[rgb]{0.9,0.95,1}\makecell{ 94.7\\81.1 } 
    & \cellcolor[rgb]{0.9,0.95,1}\makecell{ 93.6\\58.3 } 
    & \cellcolor[rgb]{0.9,0.95,1}\makecell{ 97.6\\95.1 } 
    & \cellcolor[rgb]{0.9,0.95,1}\makecell{ 77.3\\42.9 } 
    & \cellcolor[rgb]{0.9,0.95,1}\makecell{ \textbf{66.3}\\44.8 } 
    & \cellcolor[rgb]{0.9,0.95,1}\textbf{30.2}   \\

    \rowcolor[HTML]{EEE5F4}\hline
     \multicolumn{15}{c}{\textit{{Diffusion-based Planner}\quad \quad}} \\
    
    \multirow{6}{*}{{\makecell[l]{
    DiffusionDrive~\cite{liao2025diffusiondrive}}} }
    & \multirow{6}{*}{ResNet34} 
    & \textit{w/o} 
    & \makecell{S 1 \\ S 2}  
    & \makecell{ 96.8\\80.1 } 
    & \makecell{ 86.0\\72.8 } 
    & \makecell{ 98.8\\84.4 } 
    & \makecell{ 99.3\\98.4 }  
    & \makecell{ 84.0\\85.9 } 
    & \makecell{ 95.8\\76.6 } 
    & \makecell{ 96.7\\46.4 } 
    & \makecell{ 97.6\\96.3 } 
    & \makecell{ 79.6\\72.8 } 
    & \makecell{ 66.7\\40.5 } 
    & 27.5   \\
    \cmidrule{3-15}

    & 
    & \textit{recovery}  
    & \makecell{S 1 \\ S 2} 
    & \makecell{ 97.2\\82.4 } 
    & \makecell{ 88.4\\67.7 } 
    & \makecell{ 99.1\\89.1 } 
    & \makecell{ 99.8\\98.6 }  
    & \makecell{ 83.9\\89.0 } 
    & \makecell{ 96.0\\77.6 } 
    & \makecell{ 96.7\\53.8 } 
    & \makecell{ 97.6\\95.2 } 
    & \makecell{ 76.9\\46.8 } 
    & \makecell{ \textbf{69.4}\\41.7 } 
    & 30.4   \\
    \cmidrule{3-15}
    
    &  & \cellcolor[rgb]{0.9,0.95,1}\textit{\textbf{planner}} 
    & \cellcolor[rgb]{0.9,0.95,1}\makecell{S 1 \\ S 2}  
    & \cellcolor[rgb]{0.9,0.95,1}\makecell{ 97.2\\82.4 } 
    & \cellcolor[rgb]{0.9,0.95,1}\makecell{ 88.7\\72.1 } 
    & \cellcolor[rgb]{0.9,0.95,1}\makecell{ 99.3\\92.9 } 
    & \cellcolor[rgb]{0.9,0.95,1}\makecell{ 99.3\\98.5 }  
    & \cellcolor[rgb]{0.9,0.95,1}\makecell{ 82.8\\92.1 } 
    & \cellcolor[rgb]{0.9,0.95,1}\makecell{ 96.9\\80.6 } 
    & \cellcolor[rgb]{0.9,0.95,1}\makecell{ 98.0\\60.8 } 
    & \cellcolor[rgb]{0.9,0.95,1}\makecell{ 97.3\\95.4 } 
    & \cellcolor[rgb]{0.9,0.95,1}\makecell{ 59.6\\31.9 } 
    & \cellcolor[rgb]{0.9,0.95,1}\makecell{ 67.5\\ \textbf{46.8} } 
    & \cellcolor[rgb]{0.9,0.95,1}\textbf{32.6}   \\

    \rowcolor[HTML]{EEE5F4}\hline
     \multicolumn{15}{c}{\textit{{Scoring-based Planner}\quad \quad\quad }} \\
    
    \multirowcell{16}[-4ex][l]{
    GTRS-Dense~\cite{li2025gtrs}} 
    
    & \multirowcell{8}[-1.2ex]{ResNet34} 
    & \textit{w/o} 
    & \makecell{S 1 \\ S 2}  
    & \makecell{ 99.3 \\ 92.8 } 
    & \makecell{ 96.6 \\ 88.6} 
    & \makecell{ 99.6 \\ 95.5 }
    & \makecell{ 100  \\ 99.4} 
    & \makecell{ 57.4 \\ 55.9} 
    & \makecell{ 99.5 \\ 91.3} 
    & \makecell{ 92.6 \\ 55.7} 
    & \makecell{ 89.5 \\ 91.1} 
    & \makecell{16.4 \\ 35.7} 
    & \makecell[l]{67.1\\55.8 }
    & 38.3    \\
    \cmidrule{3-15} 

    & 
    & \textit{recovery}  
    & \makecell{S 1 \\ S 2}  
    & \makecell{ 97.6\\92.0 } 
    & \makecell{ 94.2\\88.1 } 
    & \makecell{ 99.3\\94.3 } 
    & \makecell{ 99.6\\98.6 }  
    & \makecell{ 76.7\\83.6 } 
    & \makecell{ 97.6\\89.8 } 
    & \makecell{ 94.9\\58.1 } 
    & \makecell{ 97.1\\91.4 } 
    & \makecell{ 37.8\\25.9 } 
    & \makecell{ 69.8\\59.3 } 
    & 43.0   \\
    \cmidrule{3-15}
    
    & 
    & \textit{planner}  
    & \makecell{S 1 \\ S 2}  
    & \makecell{ 98.7\\94.2 } 
    & \makecell{ 94.2\\92.1 } 
    & \makecell{ 99.7\\95.0 } 
    & \makecell{ 100\\99.0 }  
    & \makecell{ 74.3\\79.9 } 
    & \makecell{ 98.2\\91.8 } 
    & \makecell{ 95.8\\58.9 } 
    & \makecell{ 96.7\\91.0 } 
    & \makecell{ 34.7\\32.9 } 
    & \makecell{ 70.8\\\textbf{63.9} } 
    & 46.1   \\
    \cmidrule{3-15}
    
    &  
    & \cellcolor[rgb]{0.9,0.95,1}\textit{\textbf{reward}}
    & \cellcolor[rgb]{0.9,0.95,1}\makecell{S 1 \\ S 2} 
    & \cellcolor[rgb]{0.9,0.95,1}\makecell{ 97.6\\94.3 } 
    & \cellcolor[rgb]{0.9,0.95,1}\makecell{ 96.4\\92.7 } 
    & \cellcolor[rgb]{0.9,0.95,1}\makecell{ 99.3\\95.1 } 
    & \cellcolor[rgb]{0.9,0.95,1}\makecell{ 100\\99.5 }  
    & \cellcolor[rgb]{0.9,0.95,1}\makecell{ 75.7\\80.2 } 
    & \cellcolor[rgb]{0.9,0.95,1}\makecell{ 97.8\\91.5 } 
    & \cellcolor[rgb]{0.9,0.95,1}\makecell{ 93.3\\56.2 } 
    & \cellcolor[rgb]{0.9,0.95,1}\makecell{ 97.3\\90.6 } 
    & \cellcolor[rgb]{0.9,0.95,1}\makecell{ 32.9\\28.3 } 
    & \cellcolor[rgb]{0.9,0.95,1}\makecell{ \textbf{72.4}\\63.4 } 
    & \cellcolor[rgb]{0.9,0.95,1}\textbf{46.9}   \\
    \cmidrule{2-15} 
    
    & \multirowcell{8}[-1.2ex]{V2-99}
    & \textit{w/o}
    & \makecell{S 1 \\ S 2}  
    & \makecell{ 98.9 \\ 89.9} 
    & \makecell{ 94.9 \\ 90.5} 
    & \makecell{ 99.1\\ 94.1}
    & \makecell{ 100 \\ 99.3} 
    & \makecell{ 76.1\\ 77.6} 
    & \makecell{ 98.4\\ 88.5} 
    & \makecell{ 93.8 \\ 56.0} 
    & \makecell{ 94.9\\ 92.0} 
    & \makecell{ 37.8 \\ 30.2} 
    & \makecell{ 70.4\\ 58.5}
    &   41.9    \\
    \cmidrule{3-15}
    
    &
    & \textit{recovery}
    & \makecell{S 1 \\ S 2}  
    & \makecell{ 99.1\\95.0 } 
    & \makecell{ 98.2\\90.8 } 
    & \makecell{ 99.4\\94.6 } 
    & \makecell{ 100\\99.4 }  
    & \makecell{ 71.9\\75.5 } 
    & \makecell{ 99.1\\93.5 } 
    & \makecell{ 95.6\\57.4 } 
    & \makecell{ 95.8\\93.7 } 
    & \makecell{ 32.4\\36.8 } 
    & \makecell{ \textbf{73.4}\\63.1 } 
    & 46.4   \\
    \cmidrule{3-15}

    & 
    & \textit{planner}  
    & \makecell{S 1 \\ S 2}  
    & \makecell{ 99.3\\95.6 } 
    & \makecell{ 97.1\\91.9 } 
    & \makecell{ 99.9\\95.0 } 
    & \makecell{ 100\\98.9 }  
    & \makecell{ 67.2\\76.7 } 
    & \makecell{ 99.3\\93.7 } 
    & \makecell{ 94.0\\61.9 } 
    & \makecell{ 94.4\\90.8 } 
    & \makecell{ 23.6\\38.0 } 
    & \makecell{ 71.7\\\textbf{65.5} } 
    & 47.7   \\
    \cmidrule{3-15}

    & & \cellcolor[rgb]{0.9,0.95,1}\textit{\textbf{reward}}
    & \cellcolor[rgb]{0.9,0.95,1}\makecell{S 1 \\ S 2}  
    & \cellcolor[rgb]{0.9,0.95,1}\makecell{ 99.3\\94.9 } 
    & \cellcolor[rgb]{0.9,0.95,1}\makecell{ 97.6\\94.3 } 
    & \cellcolor[rgb]{0.9,0.95,1}\makecell{ 99.4\\94.5 } 
    & \cellcolor[rgb]{0.9,0.95,1}\makecell{ 100\\99.3 }  
    & \cellcolor[rgb]{0.9,0.95,1}\makecell{ 71.5\\79.0 } 
    & \cellcolor[rgb]{0.9,0.95,1}\makecell{ 99.6\\92.6 } 
    & \cellcolor[rgb]{0.9,0.95,1}\makecell{ 96.0\\57.8 } 
    & \cellcolor[rgb]{0.9,0.95,1}\makecell{ 95.8\\93.4 } 
    & \cellcolor[rgb]{0.9,0.95,1}\makecell{ 30.7\\30.9 } 
    & \cellcolor[rgb]{0.9,0.95,1}\makecell{ 72.5\\65.4 } 
    & \cellcolor[rgb]{0.9,0.95,1} \textbf{48.0}   \\

\bottomrule
\end{tabular}
}
\vspace{-7pt}
\label{tab:supp_navhard}
\end{table*}
\begin{table*}[htbp]
\centering
\caption{\mbox{\textbf{Detailed Results on \cmtt{navtest}.}  (\textit{recovery / planner}: recovery-based / planner-based expert; \textit{reward}: reward-only scoring)}}
\vspace{-0.1in}
\scriptsize
\resizebox{\textwidth}{!}{
\begin{tabular}{l | c|c |*{4}{p{0.03\textwidth}<{\centering}}|*{5}{p{0.03\textwidth}<{\centering}}|c}

    \toprule
    \textbf{Method}
    & \textbf{Backbone}
    & \textbf{Sim.}
    & \textbf{NC}↑ 
    & \textbf{DAC}↑
    & \textbf{DDC}↑
    & \textbf{TLC}↑
    & \textbf{EP}↑
    & \textbf{TTC}↑
    & \textbf{LK}↑
    & \textbf{HC}↑
    & \textbf{EC}↑ 
    & \textbf{EPDMS}↑  \\
    \midrule
    
     \makecell[l]{Human Agent}  & - & -  & 100 & 100 & 99.8  & 100 & 87.4 & 100 & 100 & 98.1 & 90.1 & 90.3    \\
    \midrule
            \rowcolor[HTML]{EEE5F4}\hline
     \multicolumn{13}{c}{\textit{{Regression-based Planner}}} \\
    
    \multirowcell{3}[-1.5ex][l]{
    LTF~\cite{kashyap2022transfuser}} 
    & \multirowcell{3}[-1.5ex][c]{ResNet34} 
    
    & \textit{w/o} 
    & 97.7
    & 94.0
    & 99.3
    & 99.8
    & 87.2 
    & 96.7 
    & 95.5 
    & 98.3 
    & 82.9 
    & 81.5   \\
    \cmidrule{3-13}

    & 
    & \textit{recovery} 
    & 97.9
    & 95.1
    & 99.5
    & 99.8
    & 87.7
    & 97.2
    & 97.2
    & 98.3
    & 87.1
    & 83.6  \\
    \cmidrule{3-13}

     &  
     & \cellcolor[rgb]{0.9,0.95,1}\textit{\textbf{planner}}
     & \cellcolor[rgb]{0.9,0.95,1}98.3
     & \cellcolor[rgb]{0.9,0.95,1}95.6
     & \cellcolor[rgb]{0.9,0.95,1}99.6
     & \cellcolor[rgb]{0.9,0.95,1}99.8
     & \cellcolor[rgb]{0.9,0.95,1}87.1
     & \cellcolor[rgb]{0.9,0.95,1}97.5
     & \cellcolor[rgb]{0.9,0.95,1}97.2
     & \cellcolor[rgb]{0.9,0.95,1}98.3
     & \cellcolor[rgb]{0.9,0.95,1}88.2
     & \cellcolor[rgb]{0.9,0.95,1}\textbf{84.4} \\
            \rowcolor[HTML]{EEE5F4}\hline
     \multicolumn{13}{c}{\textit{{Diffusion-based Planner\quad \quad}}} \\
   \multirowcell{3}[-1.5ex][l]{
   DiffusionDrive~\cite{liao2025diffusiondrive}} & \multirowcell{3}[-1.5ex][c]{ResNet34} 
    & \textit{w/o} 
    & 98.4
    & 95.5
    & 99.5 
    & 99.8 
    & 87.5 
    & 97.5 
    & 96.9
    & 98.4
    & 87.7
    & 84.2    \\
    \cmidrule{3-13}

    &    
    & \textit{recovery} 
    & 98.4
    & 96.7
    & 99.6
    & 99.8
    & 87.6
    & 97.5
    & 97.5
    & 98.3
    & 87.1
    & 85.4  \\
    \cmidrule{3-13}

     &  & \cellcolor[rgb]{0.9,0.95,1}\textit{\textbf{planner}}
     & \cellcolor[rgb]{0.9,0.95,1}98.5
     & \cellcolor[rgb]{0.9,0.95,1}97.1
     & \cellcolor[rgb]{0.9,0.95,1}99.6
     & \cellcolor[rgb]{0.9,0.95,1}99.8
     & \cellcolor[rgb]{0.9,0.95,1}87.4
     & \cellcolor[rgb]{0.9,0.95,1}97.8
     & \cellcolor[rgb]{0.9,0.95,1}98.0
     & \cellcolor[rgb]{0.9,0.95,1}98.3
     & \cellcolor[rgb]{0.9,0.95,1}87.5
     & \cellcolor[rgb]{0.9,0.95,1}\textbf{85.9} \\

    \rowcolor[HTML]{EEE5F4}\hline
     \multicolumn{13}{c}{\textit{{Scoring-based Planner\quad \quad\quad }}} \\

    \multirowcell{8}[-5.5ex][l]{
    GTRS-Dense~\cite{li2025gtrs}} & 
    \multirowcell{4}[-2.5ex][c]{ResNet34} 
    & \textit{w/o} 
    & 97.6
    & 97.5
    & 99.0
    & 99.9
    & 87.9 
    & 97.0 
    & 95.9 
    & 97.5
    & 55.9 
    & 82.3    \\
    \cmidrule{3-13}
    &
    & \textit{recovery} 
    & 98.2
    & 97.9
    & 99.5
    & 99.9
    & 87.2
    & 97.8
    & 96.3
    & 97.6
    & 56.5
    & 83.4  \\
    \cmidrule{3-13}

    &
    & \textit{planner} 
    & 97.8
    & 98.3
    & 99.5
    & 99.9
    & 89.2
    & 97.3
    & 97.3
    & 96.9
    & 55.1
    & 84.0  \\
    \cmidrule{3-13}

      &  & \cellcolor[rgb]{0.9,0.95,1}\textit{\textbf{reward}}
     & \cellcolor[rgb]{0.9,0.95,1}98.4
     & \cellcolor[rgb]{0.9,0.95,1}98.8
     & \cellcolor[rgb]{0.9,0.95,1}99.4
     & \cellcolor[rgb]{0.9,0.95,1}99.9
     & \cellcolor[rgb]{0.9,0.95,1}88.0
     & \cellcolor[rgb]{0.9,0.95,1}98.0
     & \cellcolor[rgb]{0.9,0.95,1}96.5
     & \cellcolor[rgb]{0.9,0.95,1}97.6
     & \cellcolor[rgb]{0.9,0.95,1}58.0
     & \cellcolor[rgb]{0.9,0.95,1}\textbf{84.6} \\
     \cmidrule{2-13} &
    \multirowcell{4}[-2.5ex][c]{V2-99} 
    
    & \textit{w/o} 
    & 97.6 
    & 98.5
    & 99.5
    & 99.9 
    & 89.5 
    & 97.2 
    & 96.8 
    & 97.2 
    & 57.2 
    & 84.0    \\
    \cmidrule{3-13}

    &  &\textit{recovery}
    & 98.3
    & 98.7
    & 99.4
    & 99.9
    & 88.9
    & 97.9
    & 97.1
    & 97.6
    & 55.9
    & 84.5  \\
    \cmidrule{3-13}
    
    &
    & \textit{planner} 
    & 98.7
    & 99.0
    & 99.5
    & 99.8
    & 86.8
    & 98.3
    & 95.9
    & 97.9
    & 58.5
    & 84.5  \\
    \cmidrule{3-13}
    
    &
    &  \cellcolor[rgb]{0.9,0.95,1}\textit{\textbf{reward}}
     & \cellcolor[rgb]{0.9,0.95,1}97.6
     & \cellcolor[rgb]{0.9,0.95,1}98.9
     & \cellcolor[rgb]{0.9,0.95,1}99.6
     & \cellcolor[rgb]{0.9,0.95,1}99.9
     & \cellcolor[rgb]{0.9,0.95,1}89.9
     & \cellcolor[rgb]{0.9,0.95,1}97.1
     & \cellcolor[rgb]{0.9,0.95,1}97.2
     & \cellcolor[rgb]{0.9,0.95,1}97.7
     & \cellcolor[rgb]{0.9,0.95,1}58.7
     & \cellcolor[rgb]{0.9,0.95,1}\textbf{84.8} \\

\bottomrule
\end{tabular}
}
\vspace{-5pt}
\label{tab:supp_navtest}
\end{table*}

% Sec.~\ref{sec:exp_setup} {\color{cvprblue}3.1}
Tab.~\ref{tab:supp_imple} provides the detailed model and training hyperparameters that complement Sec.~\ref{sec:exp_setup} implementation protocol in the main paper. We follow the NAVSIM~\cite{Dauner2024navsim} default setup and use only the \cmtt{train logs}~\cite{karnchanachari2024nuplan} of \cmtt{navtrain} for both real-data training and sim–real co-training. All planners share identical settings across the real or sim-real training setups, and each model is trained to saturation to ensure the validity of the subsequent scaling analyses.

\boldparagraph{LTF.}~\cite{kashyap2022transfuser} A regression-based planner that directly regresses future waypoints upon fused sensory latents. It employs pretrained ResNet34~\cite{he2016deep} as image encoder and is trained for 100 epochs on 8 GPUs with a total batch size of 512 and a learning rate of $2.83\times10^{-4}$.

\boldparagraph{DiffusionDrive.}~\cite{liao2025diffusiondrive} A multi-modal, DETR-style generative planner that iteratively denoises diverse trajectories using anchor-conditioned truncated diffusion. Each anchor adaptively queries the encoded image features. It employs pretrained ResNet34~\cite{he2016deep} as image encoder and applies truncated diffusion with 20 clustered anchors. It is trained for 100 epochs on 8 GPUs with a total batch size of 512 and a learning rate of $6\times10^{-4}$.

\boldparagraph{GTRS-Dense.}~\cite{li2025gtrs} A scoring-based, DETR-style planner that ranks a dense vocabulary of clustered human trajectories using multiple supervised scoring heads. Inputs are augmented with random spatial shifts. It employs pretrained ResNet34~\cite{he2016deep} or V2-99~\cite{lee2019centermask} as image encoder with an action space of 16,384 trajectories. It is trained for 50 epochs on 32 GPUs with a total batch size of 352 and a learning rate of $4\times10^{-4}$. We use a vocabulary of 16,384 trajectories for training. At inference time, 8,192 trajectories are used for \cmtt{navhard}, while the full 16,384 trajectories are used for \cmtt{navtest}.

\section{Extended Experimental Results}
\label{sec:supp_results_ext}

\subsection{Detailed Leaderboard Results}

% ~\ref{sec:exp_leaderboard} ~\ref{tab:navhard} ~\ref{tab:navtest}
% ~{\color{cvprblue}3.2} ~{\color{cvprblue}1} ~{\color{cvprblue}2}
In the main paper Sec.~\ref{sec:exp_leaderboard}, Tab.~\ref{tab:navhard} and Tab.~\ref{tab:navtest} report only the best results in \cmtt{navhard} and \cmtt{navtest} of each planner co-training (highlighted in \tightbox{blue rows}). Supplementary Tab.~\ref{tab:supp_navhard} and Tab.~\ref{tab:supp_navtest} provide full results across different pseudo-experts and the reward-only scoring method. The best results remain highlighted in \tightbox{blue rows}, allowing direct comparison with the main paper. Planner-based experts generally outperform recovery-based experts, while scoring-based planner benefits most from reward-only method.

\subsection{Complete Qualitative Results}
\label{sec:supp_results_qual}

\begin{figure*}[t!]
    \centering
    \captionsetup{type=figure}\includegraphics[width=0.94\textwidth]{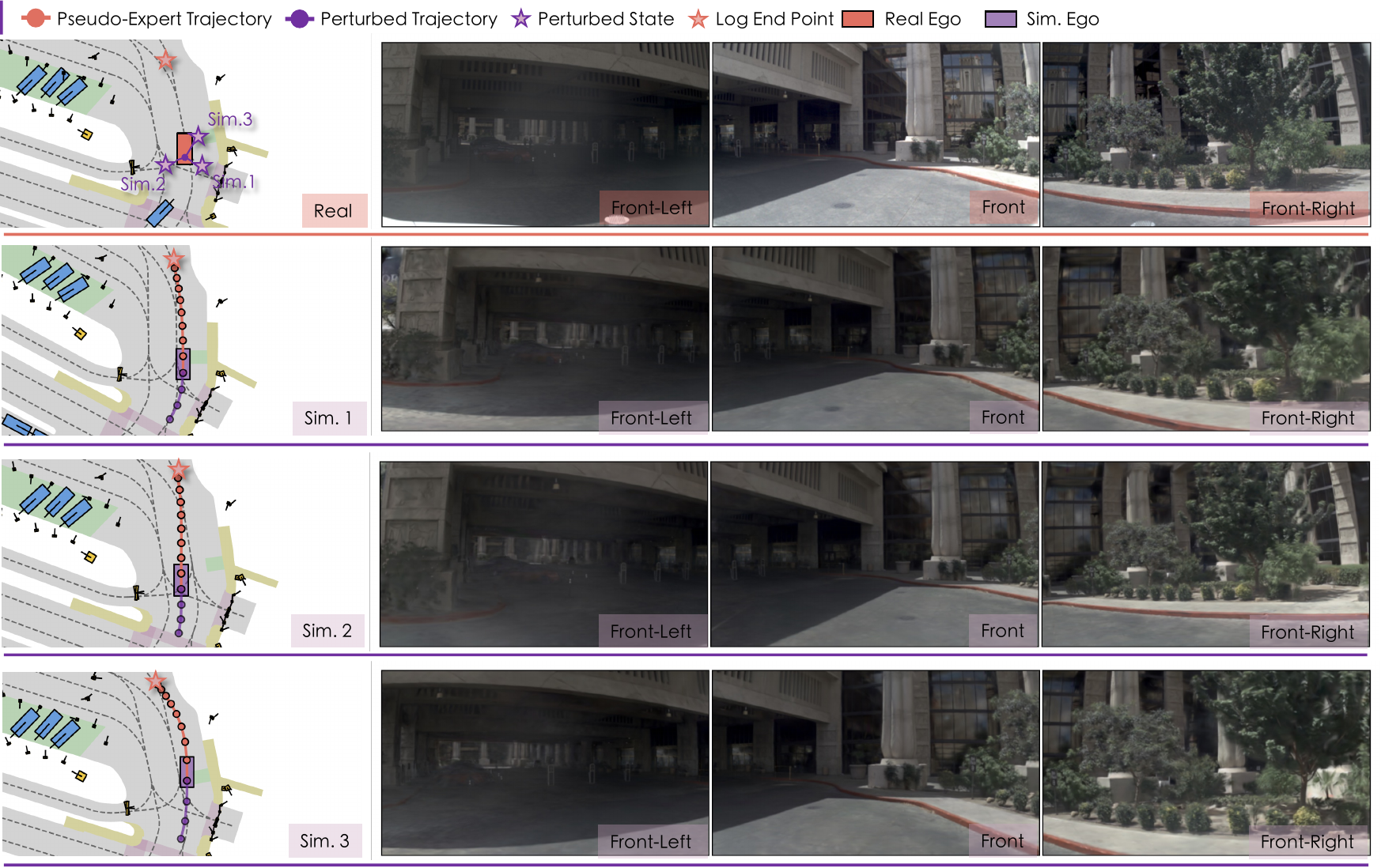} 
    \caption{\textbf{Qualitative results of recovery-based expert with real and simulation data.}}
    \label{fig:vis_expert_recovery}
    \vspace{-3pt}
\end{figure*}

\begin{figure*}[t!]
    \centering
    \captionsetup{type=figure}\includegraphics[width=0.95\textwidth]{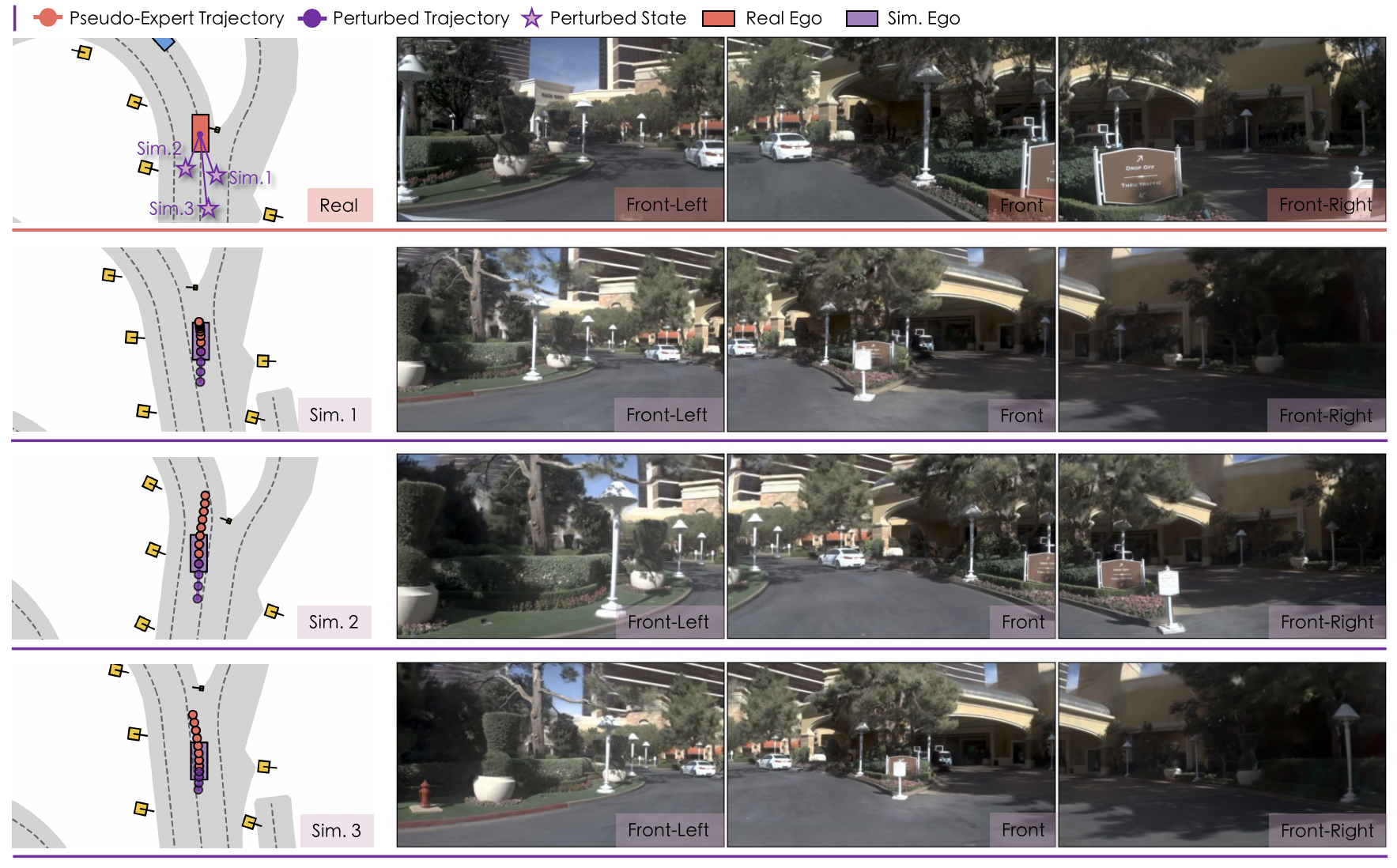} 
    \caption{\textbf{Qualitative results of planner-based expert with real and simulation data. (Y-branch intersection)} }
    \label{fig:vis_expert_planner_1}
    \vspace{-3pt}
\end{figure*}

\begin{figure*}[t!]
    \centering
    \captionsetup{type=figure}\includegraphics[width=0.95\textwidth]{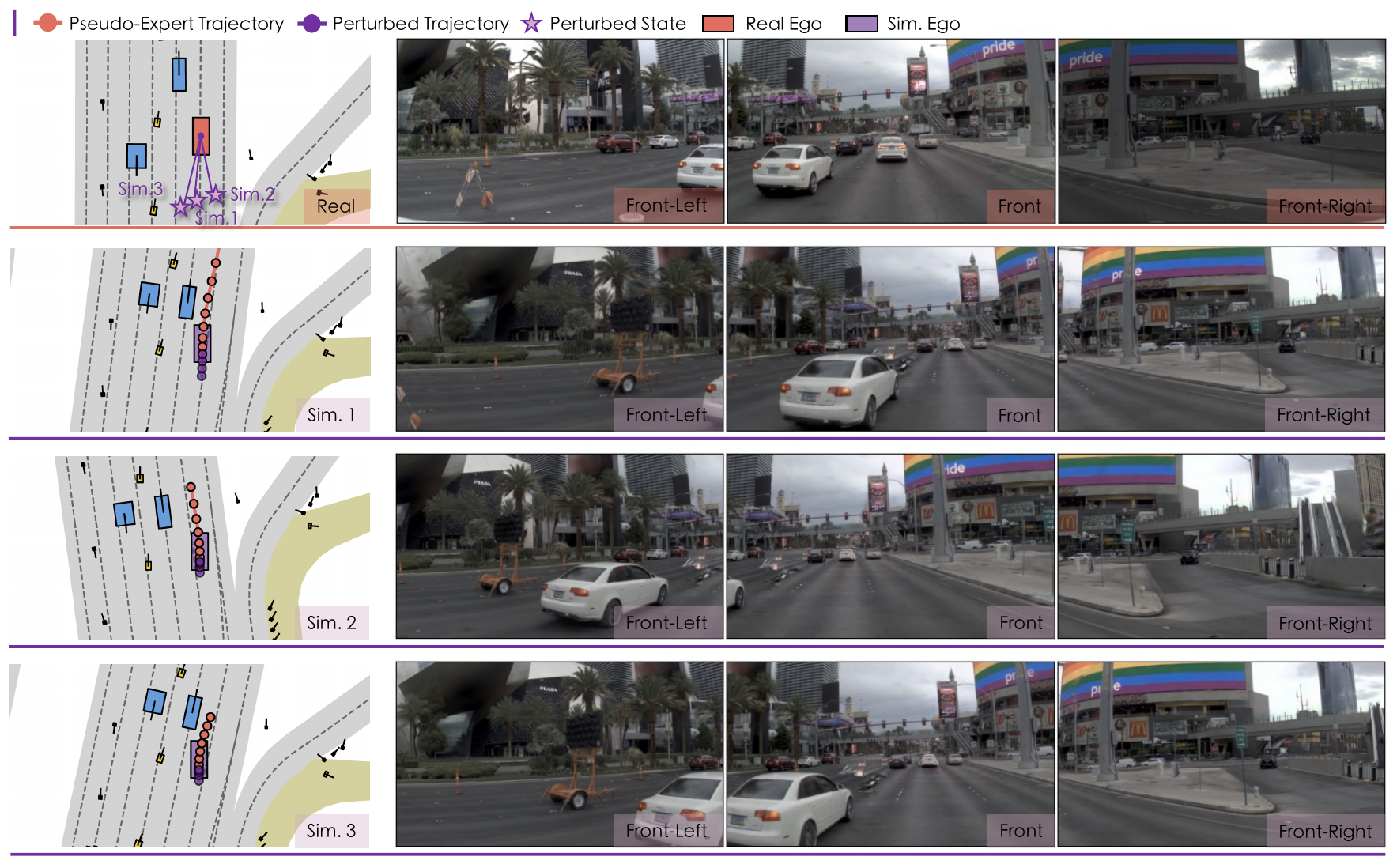} 
    \caption{\textbf{Qualitative results of planner-based expert with real and simulation data. (dense urban avenue)}}
    \label{fig:vis_expert_planner_2}
    \vspace{-3pt}
\end{figure*}

\begin{figure*}[t!]
    \centering
    \captionsetup{type=figure}\includegraphics[width=0.95\textwidth]{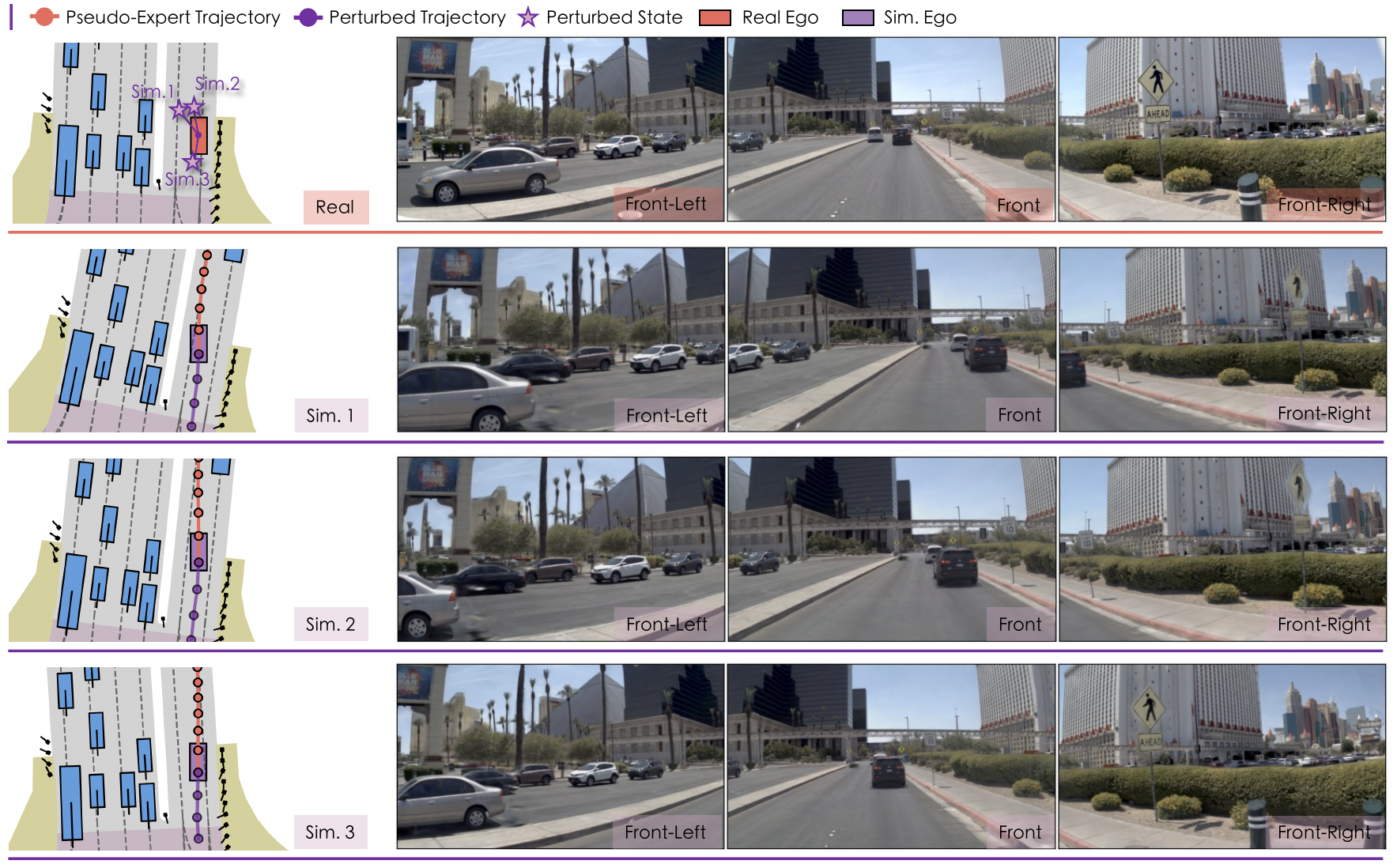} 
    \caption{\textbf{Qualitative results of planner-based expert with real and simulation data. (narrow local road)}}
    \label{fig:vis_expert_planner_3}
    \vspace{-3pt}
\end{figure*}

\begin{figure*}[t!]
    \centering
    \captionsetup{type=figure}\includegraphics[width=0.95\textwidth]{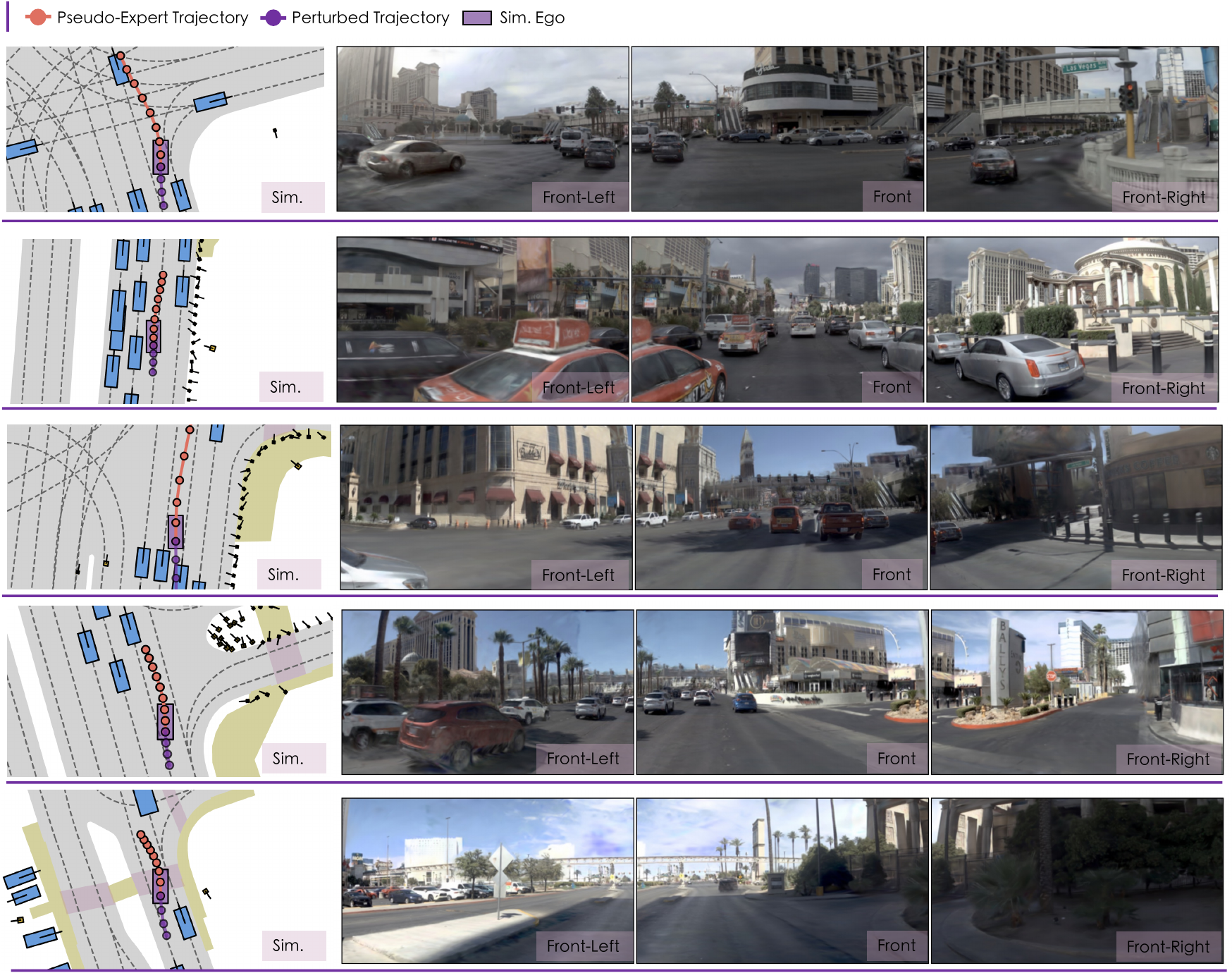} 
    \caption{\textbf{Additional qualitative results of the simulation scenes on \cmtt{navtrain}}. 
    }
    \label{fig:scene_add}
    \vspace{-3pt}
\end{figure*}

We show extended visualizations of our simulated data in Fig.~\ref{fig:vis_expert_recovery} and Figs.~\ref{fig:vis_expert_planner_1}--\ref{fig:vis_expert_planner_3}, which demonstrate the high fidelity and pseudo-expert generation of our data generation pipeline. More OOD simulation scenes are shown in Fig~\ref{fig:scene_add}.

\boldparagraph{Recovery-based Expert.} As shown in Fig.~\ref{fig:vis_expert_recovery}, all simulation trajectories successfully converge back to the human expert’s final waypoint.

\boldparagraph{Planner-based Expert} 
As shown in Figs.~\ref{fig:vis_expert_planner_1}--\ref{fig:vis_expert_planner_3}, 
the simulated feasible trajectories under perturbed states exhibit substantial variation and broader coverage across diverse traffic scenarios, including the Y-branch intersection, the dense urban avenue, and the narrow local road.

\subsection{Scaling of Reactive vs. Non-Reactive}
\label{sec:supp_results_react}
\begin{figure}[t!]
    \vspace{-5pt}
    \centering
    \hspace*{-0.05\linewidth}\includegraphics[width=1.05\linewidth]{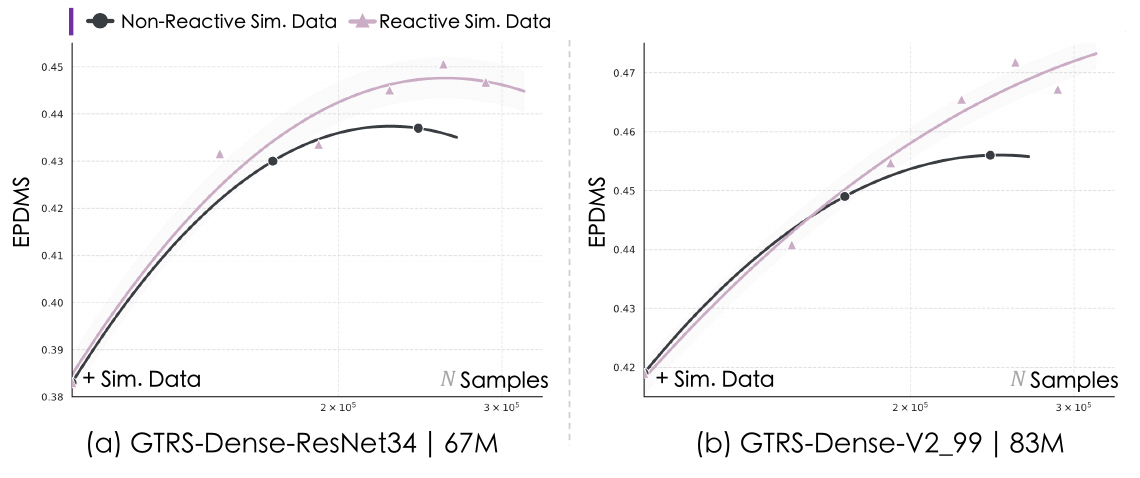}
    \caption
    {\textbf{Scaling dynamics under reactive and non-reactive simulation} using GTRS-Dense across model sizes.
    }
    \label{fig:scaling_react}
    \vspace{-10pt}
\end{figure}

In the main paper, Tab.~{\color{cvprblue}4} reports only a subset of simulation rounds for GTRS-Dense. 
Supplementary Fig.~\ref{fig:scaling_react} provides the full scaling curves, demonstrating that reactive simulation data consistently achieves superior scaling performance across model sizes.

\begin{figure*}[t]
    \vspace{-10pt}
    \centering
    % \captionsetup{skip=0.5pt}  % 控制图片和 caption 之间的距离
    \includegraphics[width=0.95\linewidth]{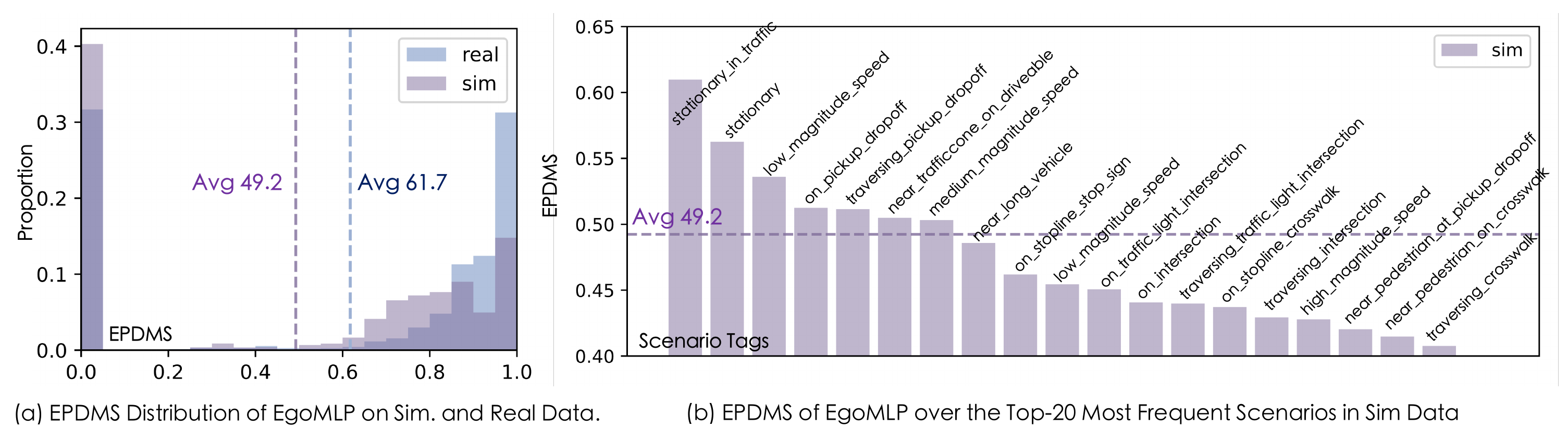}
    \caption{\textbf{EgoMLP $\mathrm{EPDMS}$ on simulation and real data.} 
    (a) $\mathrm{EPDMS}$ distribution under simulation versus real-world data. 
    (b) Scenario tags ranked by $\mathrm{EPDMS}$. 
    The figures reveal the distribution shift and scenario characteristics of simulation data.}
    \label{fig:data_statistic}
    \vspace{-14pt}
\end{figure*}

\section{Additional Experimental Results}
\label{sec:supp_results_add}

\subsection{Multi-Expert Ensemble}
\begin{table}[H]
\vspace{-5pt}
\centering
\caption{\textbf{The effect of multi-expert ensemble} on \cmtt{navhard} using GTRS-Dense. ($\textbf{S}_{1/2}$:Per-stage EPDM scores; \textit{recovery / planner}: recovery-based / planner-based expert; \textit{reward}: reward scoring only; $*$:multi-expert ensemble)}
\small
\setlength{\tabcolsep}{6.5pt}
\begin{tabular}{c|cc|ccc}
\toprule
\textbf{Backbone} 
& \textbf{Sim} 
& \textbf{Expert} 
& $\textbf{S}_1 \uparrow$
& $\textbf{S}_2 \uparrow$
& $\textbf{EPDMS} \uparrow$ 
\\
\midrule
\multirowcell{5}[-1ex]{ResNet34}
& 
& 
&67.1 
&55.8 
&38.3  \\
\cmidrule{2-6}

& \checkmark   
& \textit{recovery}
&69.8
&59.3
&43.0 \\

& \checkmark   
& \textit{planner}
&70.8
&63.9
&46.1 \\

& \checkmark   
&\textit{reward}
&72.4
&63.4
&46.9\\

\cmidrule{2-6}

& \cellcolor[rgb]{0.9,0.95,1}\checkmark   
& \cellcolor[rgb]{0.9,0.95,1}$*$
&\cellcolor[rgb]{0.9,0.95,1}\textbf{72.8}
&\cellcolor[rgb]{0.9,0.95,1}\textbf{65.1}
&\cellcolor[rgb]{0.9,0.95,1}\textbf{47.7}\\

\midrule

\multirowcell{5}[-1ex]{V2-99}
& 
& 
&70.4 
& 58.5 
& 41.9  \\
\cmidrule{2-6}
 
& \checkmark   
& \textit{recovery}
&73.4
&63.1
&46.4 \\

& \checkmark   
& \textit{planner}
&71.7
&65.5
&47.7 \\

& \checkmark   
&\textit{reward}
&72.5
&65.4
&48.0 \\

\cmidrule{2-6}

& \cellcolor[rgb]{0.9,0.95,1}\checkmark   
& \cellcolor[rgb]{0.9,0.95,1}$*$
&\cellcolor[rgb]{0.9,0.95,1}\textbf{74.8}
&\cellcolor[rgb]{0.9,0.95,1}\textbf{67.6}
&\cellcolor[rgb]{0.9,0.95,1}\textbf{50.9}\\

\midrule

\textcolor[rgb]{0.5,0.5,0.5}{$*$}
& \textcolor[rgb]{0.5,0.5,0.5}\checkmark  
& \textcolor[rgb]{0.5,0.5,0.5}{$*$}
& \textcolor[rgb]{0.5,0.5,0.5}{75.4}
& \textcolor[rgb]{0.5,0.5,0.5}{66.4}
& \textcolor[rgb]{0.5,0.5,0.5}{50.4} \\

\bottomrule
\end{tabular}
\vspace{-1.5pt}
\label{tab:supp_ens}
\end{table}

Different pseudo-experts exhibit distinct behavioral characteristics, \eg, the recovery-based expert is more conservative, whereas the planner-based expert is more exploratory. This raises the question of whether these complementary properties can be leveraged jointly. Therefore, we conduct a simple ensemble study, multi-expert enseble as we call it, to examine their potential complementarity on \cmtt{navhard}, as shown in Tab.~\ref{tab:supp_ens}. The scoring-based GTRS-Dense planner is chosen because it enables a straightforward ensemble: the predicted sub-scores of each trajectory in the vocabulary can be directly averaged across the three models (\textit{recovery-based}, \textit{planner-based}, \textit{reward-only scoring}).

\boldparagraph{Ensemble Gains from Expert Diversity.} As shown in Tab.~\ref{tab:supp_ens}, this simple multi-expert ensemble yields consistent improvements for both backbones, achieving +0.8 and +2.9 $\mathrm{EPDMS}$ on ResNet34 and V2-99 compared with reward-only scoring, respectively, for \cmtt{navhard}. Further enlarging the ensemble to all six models yields no additional $\mathrm{EPDMS}$ gains.
These indicate that, for end-to-end planning, different pseudo-experts provide strong complementary benefits—often exceeding the gains brought by different backbone structures.

\subsection{Effect of Scenarios beyond Scaling}

Beyond data scaling, simulation data exhibits an out-of-distribution shift relative to real-world data.
We provide a preliminary analysis of these characteristics by evaluating an EgoMLP planner, trained on \cmtt{navtrain}, on both \cmtt{navtest} and simulation data, as shown in Fig.~\ref{fig:data_statistic}.
(1) Fig.~\ref{fig:data_statistic} (a): Sim data shows a lower $\mathrm{EPDMS}$ distribution than real data, confirming a higher safety-critical level.
(2) Fig.~\ref{fig:data_statistic} (b): Low $\mathrm{EPDMS}$ scores across nuPlan scenario tags identify challenging simulation scenarios, facilitating future study on their distinct impact, \eg, increasing their proportions during simulation. 

\subsection{Effect of Simulation Visual Fidelity}
\begin{table}[H]
    \vspace{-7pt}
    \centering
    \caption{\textbf{Effect of simulation visual fidelity} on \cmtt{navhard} using GTRS-Dense, across sampling rounds and co-training strategies.}
    \resizebox{1\linewidth}{!}{
    \begin{tabular}{c| c | c c | c c}
    \toprule
    \multirow{2}{*}{\textbf{Backbone}} &
    \multirow{2}{*}{\textbf{$\mathrm{PSNR}$}} &
    \multicolumn{2}{c|}{\footnotesize\textbf{Pseudo-Expert}} &
    \multicolumn{2}{c}{\footnotesize\textbf{Reward-Only Scoring}} \\
    \cmidrule(lr){3-4} \cmidrule(lr){5-6}
    & &
    \textbf{\footnotesize \#R 1} & 
    \textbf{\footnotesize \#R 2} & 
    \textbf{\footnotesize \#R 1} & 
    \textbf{\footnotesize \#R 2} \\
    \midrule
    \multirow{2}{*}{ResNet34} 
    & $<27$ & 41.0 & 41.1 & 40.8 & 41.6 \\
    & $\geq27$ & \textbf{41.9} & \textbf{41.2} & \textbf{41.3} & \textbf{42.9} \\
     \midrule
    \multirow{2}{*}{V2-99} 
    & $<27$ & 44.4 & 44.5 & 43.7 & 44.8 \\
    & $\geq27$ & \textbf{45.4} & \textbf{46.1} & \textbf{44.9} & \textbf{46.3} \\
    \bottomrule
    \end{tabular}
    }
    \vspace{-10pt}
    \label{tab:psnr}
\end{table}

We retrain GTRS-Dense with new simulation data of $\mathrm{PSNR}<27$ ($\sim$10K$\times$2 rounds) and compare it with the matched samples randomly selected from simulations with $\mathrm{PSNR}\geq27$. The former represents lower reconstruction quality and is more likely to yield simulations with reduced visual fidelity. As shown in Tab.~\ref{tab:psnr}, simulation data with $\mathrm{PSNR}\geq27$ consistently yields higher EPDMS under different data scales and co-training strategies. These results indicate that simulation data with higher visual fidelity helps reduce the visual gap between simulation and real-world observations, thereby improving the effectiveness of simulation data.

\subsection{Scaling with Varying Scales of Real Data}

\begin{figure}[H]
    \vspace{-3pt}
    \centering
    \hspace*{-0.05\linewidth}\includegraphics[width=1.05\linewidth]{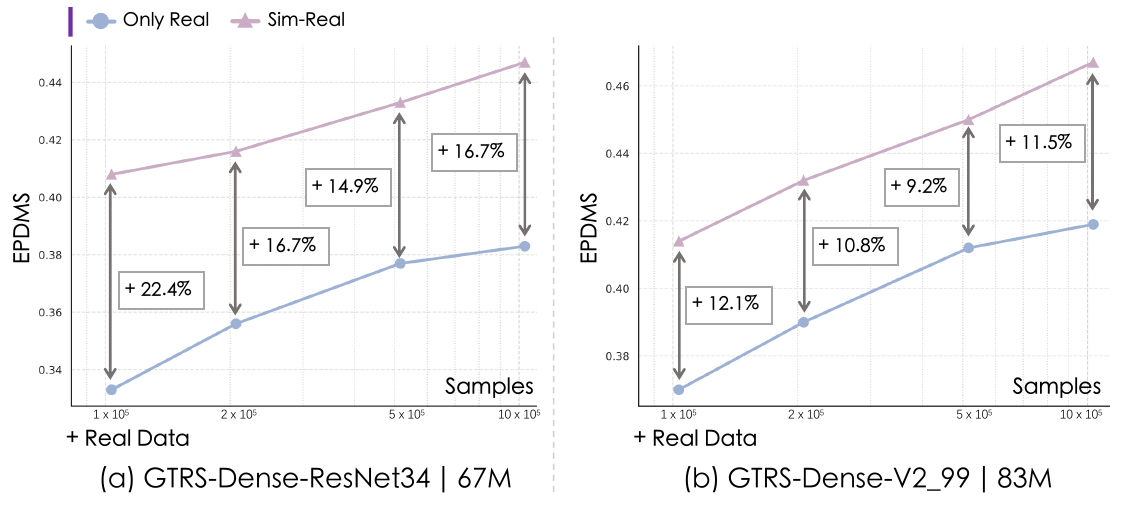}
    \caption
    {\textbf{Scaling simulation with varying real data.} Simulation data are scaled by corresponding real data scenario tokens and fixed sim-real data ratio
    }
    \label{fig:supp_scaling_real}
    \vspace{-3pt}
\end{figure}

To assess the practical utility of our approach, we study how simulation data affects performance as the amount of available real-world data scales up. Based on GTRS-Dense with reward-only scoring sim-real co-training, we conduct experiments with different real data from \cmtt{navtrain} (10K, 20K, 50K, 100K). Inspired by ~\cite{alhaija2025cosmos-transfer1}, simulation data are generated from identical scenarios of corresponding real data tokens, and we fix the sim-real data ratio (five sampling rounds) for each experiment.

\boldparagraph{Sustained Simulation Gains Across Real Data Scales.} As shown in Fig.~\ref{fig:supp_scaling_real}, our simulation data provides the most significant gains when real data is scarce (10K), achieving +22.4\% $\mathrm{EPDMS}$ on ResNet34 and +12.1\% $\mathrm{EPDMS}$ on V2-99 for \cmtt{navhard}. As the amount of real data increases (from 10K to 100K), these gains remain consistently high, without noticeable narrowing. This demonstrates that our simulation data can effectively complement limited real-world data and still offers substantial potential to further exploit and amplify the value of real data even when it is abundant.

\section{Limitations and Broader Impact}
\label{sec:supp_limit}

\subsection{Pseudo-Expert}

Despite its effectiveness, our pseudo-expert generation pipeline has limitations. Current rule-based trajectory perturbations are static. A potential enhancement is the self-evolving approach~\cite{ye2025self-evolve}, using pretrained planners to iteratively explore the simulator and collect recovery data. Additionally, the privileged BEV planner we use is rule-based with limited performance, causing some degradation in comfort metrics ($\mathrm{HC}$, $\mathrm{EC}$ in Tab.~\ref{tab:supp_navhard}) and failing in extremely corner cases. More advanced learning-based BEV planners~\cite{zheng2025diffusionplanner,tan2025flowplanner,jaeger2025carl} could further improve generation efficiency and realism.

\subsection{Scene Simulation}

For traffic behavior simulation in our decoupled scene simulation paradigm, other agents are controlled by IDM~\cite{treiber2000idm}, which enables interaction but limits scenario diversity. Diffusion-based traffic generators~\cite{zhou2025nexus, jiang2024scenediffuser,li2025omega} offer a promising improvement. For sensor simulation, feedforward GS~\cite{charatan23pixelsplat, yang2024storm} can improve reconstruction efficiency, while adding modalities like LiDAR~\cite{li2025uniscene} provides complementary modality information. Furthermore, unified world models~\cite{wu2025umgen, yang2025unifiedsurvey, zhang2025scaling, zhang2025latent, shi2025realunify} can serve as a substitute for both behavior and sensor simulation.

\subsection{Self-Play}

Self-play~\cite{cusumano2025selfplay, cornelisse2025simagent} allows the ego and surrounding agents to share learned policies instead of relying on separate traffic behavior simulators, enabling behaviors to co-evolve through interaction. With sensor simulation, it can support end-to-end policy learning and potentially improve robustness in long-tail scenarios, though it remains limited by interaction cost and simulation fidelity.

\subsection{Societal Impact}

Despite promising improvements, pseudo-expert scene simulation still has room for quality and efficiency enhancements for real-world deployment. Co-training may be affected by unrealistic simulation visuals and distributional differences in real-world corner cases, which could lead to potential risks. We hope \name~will inspire both academia and industry in driving and robotics to leverage real-world simulation to address rare corner cases in data scaling, advancing fully autonomous systems that are robust and generalizable. In addition, we will provide the community with a fully open-source simulation dataset and training framework to promote academic research on autonomous driving in simulation.

\section{License of Assets}

All training and evaluation are performed on data from publicly licensed datasets~\cite{karnchanachari2024nuplan, openscene2023, Cao2025navsimv2}. The real-world data engine is based on the MTGS codebase~\cite{li2025mtgs} under the Apache-2.0 license. We adopt publicly available end-to-end planners, including GTRS~\cite{li2025gtrs} (Apache-2.0) and DiffusionDrive~\cite{liao2025diffusiondrive} and LTF~\cite{kashyap2022transfuser} (MIT).

\label{sec:supp_license}


\end{document}